%% file: main.tex
\PassOptionsToPackage{unicode}{hyperref}
\PassOptionsToPackage{hyphens}{url}
\PassOptionsToPackage{dvipsnames,svgnames,x11names}{xcolor}
\documentclass[
  12pt]{article}

\input{macros}
\usepackage{booktabs}
\usepackage{tabularx}
\usepackage{needspace}
\usepackage{multirow}
\usepackage{amsthm,amsmath,amsfonts}

\usepackage[section]{placeins}

\theoremstyle{plain}
\newtheorem{theorem}{Theorem}[section]
\newtheorem{proposition}[theorem]{Proposition}

\theoremstyle{definition}

\newtheorem{assumption}[theorem]{Assumption}
\newtheoremstyle{italicremark}
  {\topsep}{\topsep}{\itshape}{0pt}{\itshape}{.}{5pt plus 1pt minus 1pt}{}
\theoremstyle{italicremark}
\newtheorem{remark}[theorem]{Remark}

\usepackage{amsmath,amssymb}
\usepackage{iftex}
\ifPDFTeX
  \usepackage[T1]{fontenc}
  \usepackage[utf8]{inputenc}
  \usepackage{textcomp} %
\else %
  \usepackage{unicode-math}
  \defaultfontfeatures{Scale=MatchLowercase}
  \defaultfontfeatures[\rmfamily]{Ligatures=TeX,Scale=1}
\fi
\usepackage{lmodern}
\ifPDFTeX\else  
\fi
\IfFileExists{upquote.sty}{\usepackage{upquote}}{}
\IfFileExists{microtype.sty}{%
  \usepackage[]{microtype}
  \UseMicrotypeSet[protrusion]{basicmath} %
}{}
\makeatletter
\@ifundefined{KOMAClassName}{%
  \IfFileExists{parskip.sty}{%
    \usepackage{parskip}
  }{%
    \setlength{\parindent}{0pt}
    \setlength{\parskip}{6pt plus 2pt minus 1pt}}
}{%
  \KOMAoptions{parskip=half}}
\makeatother
\usepackage{xcolor}
\makeatletter
\ifx\paragraph\undefined\else
  \let\oldparagraph\paragraph
  \renewcommand{\paragraph}{
    \@ifstar
      \xxxParagraphStar
      \xxxParagraphNoStar
  }
  \newcommand{\xxxParagraphStar}[1]{\oldparagraph*{#1}\mbox{}}
  \newcommand{\xxxParagraphNoStar}[1]{\oldparagraph{#1}\mbox{}}
\fi
\ifx\subparagraph\undefined\else
  \let\oldsubparagraph\subparagraph
  \renewcommand{\subparagraph}{
    \@ifstar
      \xxxSubParagraphStar
      \xxxSubParagraphNoStar
  }
  \newcommand{\xxxSubParagraphStar}[1]{\oldsubparagraph*{#1}\mbox{}}
  \newcommand{\xxxSubParagraphNoStar}[1]{\oldsubparagraph{#1}\mbox{}}
\fi
\makeatother

\usepackage{longtable,booktabs,array}
\usepackage{calc} %
\usepackage{etoolbox}
\makeatletter
\patchcmd\longtable{\par}{\if@noskipsec\mbox{}\fi\par}{}{}
\makeatother
\IfFileExists{footnotehyper.sty}{\usepackage{footnotehyper}}{\usepackage{footnote}}
\makesavenoteenv{longtable}
\usepackage{graphicx}
\makeatletter
\def\maxwidth{\ifdim\Gin@nat@width>\linewidth\linewidth\else\Gin@nat@width\fi}
\def\maxheight{\ifdim\Gin@nat@height>\textheight\textheight\else\Gin@nat@height\fi}
\makeatother
\setkeys{Gin}{width=\maxwidth,height=\maxheight,keepaspectratio}
\makeatletter
\def\fps@figure{htbp}
\makeatother

\makeatletter
\@ifpackageloaded{caption}{}{\usepackage{caption}}
\AtBeginDocument{%
\ifdefined\contentsname
  \renewcommand*\contentsname{Table of contents}
\else
  \newcommand\contentsname{Table of contents}
\fi
\ifdefined\listfigurename
  \renewcommand*\listfigurename{List of Figures}
\else
  \newcommand\listfigurename{List of Figures}
\fi
\ifdefined\listtablename
  \renewcommand*\listtablename{List of Tables}
\else
  \newcommand\listtablename{List of Tables}
\fi
\ifdefined\figurename
  \renewcommand*\figurename{Figure}
\else
  \newcommand\figurename{Figure}
\fi
\ifdefined\tablename
  \renewcommand*\tablename{Table}
\else
  \newcommand\tablename{Table}
\fi
}
\@ifpackageloaded{float}{}{\usepackage{float}}
\floatstyle{ruled}
\@ifundefined{c@chapter}{\newfloat{codelisting}{h}{lop}}{\newfloat{codelisting}{h}{lop}[chapter]}
\floatname{codelisting}{Listing}

\makeatother
\makeatletter
\@ifpackageloaded{caption}{}{\usepackage{caption}}
\@ifpackageloaded{subcaption}{}{\usepackage{subcaption}}
\makeatother

\ifLuaTeX
  \usepackage{selnolig}  %
\fi
\usepackage[]{natbib}
\usepackage{bookmark}

\IfFileExists{xurl.sty}{\usepackage{xurl}}{} %
\hypersetup{
  pdftitle={A Statistical Framework for Training Data Auditing of Large Language Models},
  pdfauthor={Shengjie Niu; Yeheng Ge; Jian Huang},
  pdfkeywords={Training Data Auditing, Membership Inference, Large Language Model, Multiple Testing},
  colorlinks=true,
  linkcolor={blue},
  filecolor={Maroon},
  citecolor={Blue},
  urlcolor={Blue},
  pdfcreator={LaTeX via pandoc}}

\newcommand{\anon}{1}

\begin{document}

\def\spacingset#1{\renewcommand{\baselinestretch}%
{#1}\small\normalsize} \spacingset{1}

\raggedbottom
\renewcommand{\floatpagefraction}{.9}

\if1\anon
{
  \title{\bf A  Statistical Framework for Training Data Auditing of Large Language Models}
  \author{\small Shengjie $\rm{Niu}^{a \ *}$, Yeheng $\rm{Ge}^{a}$
  \thanks{Co-first authors.},
  Jian $\rm{Huang}^{a,b}$\thanks{Corresponding author. }\\
    {\footnotesize $\it^{a}$Department of Data Science and Artificial Intelligence, The Hong Kong Polytechnic University}\\
    {\footnotesize $\it^{b}$Department of Applied Mathematics, The Hong Kong Polytechnic University}\\
    }
  \maketitle
} \fi

\if0\anon
{
  \bigskip
  \bigskip
  \bigskip
  \begin{center}
    {\LARGE\bf A Statistical Framework for Training Data Auditing of Large Language Models}
\end{center}
  \medskip
} \fi

\bigskip

\input{0_abstract}

\noindent%
{\it Keywords:}
Membership inference;
multiple testing;
large language model;
training data auditing.
\vfill

\newpage
\spacingset{1.8} %

\input{1_intro}
\input{3_problem}
\input{4_method}

\input{5_theorey}
\input{6_exp}
\input{7_conclusion}

{ \spacingset{1.5}
\bibliography{bibliography.bib}
}

\end{document}


\def\spacingset#1{\renewcommand{\baselinestretch}%
{#1}\small\normalsize} \spacingset{1}

\if1\anon
{
  \title{\bf Title}
  \author{Author 1\thanks{
    The authors gratefully acknowledge \textit{please remember to list all relevant funding sources in the version that gives all author information}}\hspace{.2cm}\\
    Department of YYY, University of XXX\\
    and \\
    Author 2 \\
    Department of ZZZ, University of WWW}
  \maketitle
} \fi

\if0\anon
{
  \bigskip
  \bigskip
  \bigskip
  \begin{center}
    {\LARGE\bf Supplementary Material of ``A Statistical Framework for Training Data Auditing of Large Language Models''}
\end{center}
  \medskip
} \fi

\bigskip

\spacingset{1.8} %

\input{X_appendix}

\bibliography{bibliography.bib}

%% file: macros.tex
\newif\ifdraft
\drafttrue

\ifdraft
  \newcommand{\todocolor}[2]{\textcolor{#1}{#2}}
\else
  \newcommand{\todocolor}[2]{}
\fi

\newcommand{\figref}[1]{Figure~\ref{#1}}

\newcommand{\secref}[1]{Section~\ref{#1}}

\newcommand{\eqnref}[1]{Equation~(\ref{#1})}

\newcommand{\mD}{\mathcal{D}}
\newcommand{\mF}{\mathcal{F}}

\newcommand{\mS}{\mathcal{S}}

\newcommand{\PP}{\mathbb{P}}

%% file: 0_abstract.tex
\begin{abstract}
Training-data auditing asks whether data of interest were used to train a large language model (LLM), a question central to privacy auditing, copyright protection, and benchmark-contamination detection.
These concerns are particularly relevant to the use of copyrighted books and scientific papers in pre-training, as well as synthetic text and medical question–answer records in fine-tuning.
In practice, training membership signals may be concentrated in only a few passages of a long document, making it important to aggregate evidence within each document without diluting these localized signals. Beyond individual documents, auditing a target collection requires an overall assessment based on the document-level evidence.
In this paper, we propose a unified statistical framework for document-level membership inference, target-collection global auditing, and suspicious-subset identification.
We integrate complementary chunk-level membership evidence with Cauchy combination to construct document scores for document-level membership inference.
The document scores and their calibrated $p$-values further enable an overall assessment of training-data collection use and the identification of suspicious documents through multiple testing.
Under the stated exchangeability conditions, we establish finite-sample validity of document-level $p$-values, type-I error control for global auditing, and false discovery rate control for subset identification.
Case studies on BookTection, Pile–arXiv, Cosmopedia, and HealthCareMagic demonstrate the effectiveness and reliability of the proposed framework in auditing pre-training and fine-tuning data.

\end{abstract}

%% file: 1_intro.tex
\section{Introduction}\label{sec:intro}

Large language models (LLMs) are trained on vast text corpora whose composition is often undisclosed~\citep{brown2020language}.
Their ability to memorize and reproduce training content raises privacy and copyright concerns~\citep{tang2024privacy-preserving,duarte2024decop}, while the inclusion of benchmark evaluation data can compromise model assessment~\citep{oren2023proving}.
These concerns motivate training-data auditing, which seeks to determine whether particular texts were used during training~\citep{mink}.
Membership inference (MIA)~\citep{mia} formulates this question as a binary decision between \emph{member} and \emph{non-member} status.
This task is fundamentally a statistical decision problem: the auditor must extract membership signals from limited and heterogeneous model outputs, aggregate diverse and dependent evidence, and translate the aggregate into membership decisions with controlled error rates.
Meeting these requirements calls for principled statistical methodology for training-data auditing at both the document and collection levels.
Accordingly, we develop a unified statistical framework for training-data auditing that addresses three complementary objectives: document-level membership inference, global auditing of a target collection, and identification of suspicious subsets.

The first objective is \textbf{document-level auditing}: determining whether a specific document was included in the target model’s training set.
Documents such as scientific papers and Wikipedia pages can span thousands of tokens, while membership signals may be concentrated in localized paragraphs~\citep{wbc}.
Auditing such documents therefore involves extracting evidence from smaller units, referred to as \emph{chunks}, and aggregating it at the document level~\citep{mia_scaling}.
For each chunk, auditing methods derive membership evidence statistics from model outputs, such as perplexity and log-probability-based scores. These statistics exploit differences in model fit: the overfitting nature of large language models can lead to lower prediction loss on training text than on non-member text, corresponding to lower perplexity and higher average token log-probabilities~\citep{maini2024datasetllm}. These statistics may be dependent across chunks within the same document, making evidence combination and aggregation statistically challenging.

The second objective is \textbf{target-collection global auditing}: examining evidence of training-data use across a collection of documents.
\emph{Dataset inference}~\citep{maini2024datasetllm} addresses this objective by testing whether any members of the target dataset were used for training.
This problem is particularly relevant to data owners, businesses, and rights holders seeking to verify large-scale use of their data.
For example, a scholarly publisher may suspect that its journal corpus was used in LLM pre-training, while an AI company may seek to determine whether its proprietary reasoning dataset was used to fine-tune a competitor’s model.
However, the statistical tests and uncertainty estimates used by these approaches rely on independence assumptions and lack corresponding validity guarantees for collections of dependent documents.

The third objective is \textbf{suspicious-subset identification}. It determines which specific documents in a target collection are likely to be training members. Existing dataset-level methods mostly provide aggregate conclusions about collection-level usage and have paid comparatively little attention to itemized evidence.
In copyright and compliance auditing, rights holders and auditors often need an itemized list of suspicious documents to support data-use claims and guide further verification. Because such a list may support legal or regulatory action, the proportion of false inclusions must be controlled.
We formalize this requirement through false discovery rate (FDR) control under dependent document signals~\citep{benjamini1995controlling}.
The result is an itemized list in which the expected proportion of false accusations is bounded by a prespecified level $q$, yielding evidence reliable enough to support legal, regulatory, or organizational review.

In this paper, we propose a multi-level statistical framework for training-data auditing, as illustrated in \figref{fig:teaser}.
The framework unifies the three auditing objectives by translating chunk-level membership evidence into document-level conclusions and collection-level decisions.
These transitions introduce distinct statistical challenges.
Within a document, heterogeneous and dependent chunk-level statistics complicate aggregation, which may dilute localized membership signals or amplify noisy evidence~\citep{wbc}.
Across documents, the signals are also dependent because all documents are scored by the same fitted rule and calibrated against the same non-member collection.
This dependence among document scores complicates both the calibration of global tests and FDR control for suspicious-subset identification.
To address these challenges, we integrate complementary statistical tools into a unified auditing framework: Cauchy combination~\citep{liu2020cauchy} aggregates local evidence into document scores sensitive to localized signals, and empirical calibration against held-out non-member documents then provides a basis for individual membership testing. 
At the collection level, an Energy permutation test~\citep{szekely2013energy} is applied to assess distributional departures from non-member behavior, while the Benjamini–Hochberg (BH) procedure~\citep{benjamini1995controlling} is considered to select suspicious documents using the calibrated document-level $p$-values. 
Under the stated exchangeability conditions, which permit dependence among document scores, we establish type-I error control for individual and global testing and FDR control for subset identification.

Our framework applies to both pre-training and fine-tuning data auditing. 
We study four case studies to evaluate the proposed framework. For pre-training, we examine copyrighted books from BookTection and scientific papers from Pile–arXiv, where auditing involves aggregating local membership evidence across long documents. For fine-tuning, we examine synthetic text from Cosmopedia and medical question–answer records from HealthCareMagic to assess whether these data were used to fine-tune the pretrained models. Across these case studies, we assess the framework’s detection performance and statistical reliability at all the document, collection, and subset levels.

Our contributions are threefold.
First, we formulate LLM training-data auditing through three complementary statistical objectives: document-level membership inference, target-collection global auditing, and suspicious-subset identification with explicit error control.
Second, we develop a unified framework that combines heterogeneous chunk-level evidence and uses Cauchy combination to construct document scores sensitive to localized membership signals. These scores support empirically calibrated individual testing, Energy-based global auditing, and BH subset identification. 
Third, under the stated conditional exchangeability conditions, we establish finite-sample validity of document-level calibration, type-I error control for individual and global testing, and FDR control for subset identification. We further quantify the additional error caused by departures from exchangeability.

The remainder of this paper is organized as follows.
\secref{sec:problem} introduces the preliminaries, the datasets, and formulates the auditing problems.
\secref{sec:method} presents the proposed statistical auditing framework.
\secref{sec:theoretical_guarantees} establishes its theoretical guarantees under conditional exchangeability and quantifies the effects of departures from these conditions.
\secref{sec:exp} evaluates the framework through four case studies spanning pre-training and fine-tuning data auditing.
\secref{sec:conclusion} concludes the paper.

%% file: 3_problem.tex
\section{Preliminaries, Datasets, and Problems}\label{sec:problem}

\subsection{Preliminaries}\label{sec:problem_preliminaries}

\paragraph*{Autoregressive Language Models}
Autoregressive LLMs model by predicting each token conditional on its preceding context~\citep{brown2020language}.
Let $\mathbf{w}=(w_1,\ldots,w_L)$ denote a token sequence of length $L$, and let $\mathbf{w}_{<t}=(w_1,\ldots,w_{t-1})$ denote the prefix preceding token $w_t$.
For a model with parameters $\theta$, the sequence likelihood factorizes as
\begin{equation}
p_{\theta}(\mathbf{w}) = \prod_{t=1}^{L} p_{\theta}(w_t \mid \mathbf{w}_{<t}),
\end{equation}
where $p_{\theta}(w_t\mid\mathbf{w}_{<t})$ is the conditional probability assigned to token $w_t$.
These models are typically trained by minimizing the negative log-likelihood (NLL) of the training sequences.

These token-level conditional probabilities are also the quantities on which auditing is based. Evaluating them for a given sequence requires a single forward pass rather than text generation, and their negative logarithms are the per-token losses from which perplexity and the other membership evidence statistics below are obtained.

\paragraph*{Auditor Access}
We consider a \emph{black-box} setting in which the auditor can query the target model with a token sequence and obtain the conditional probability of each token given its preceding context.
The auditor has no access to the model's internal parameters or the membership labels of the target documents.
Such token probabilities can be obtained through token-scoring APIs or self-hosted model interfaces.
The Min-$K\%++$ features used in our experiments additionally require access to the full next-token predictive distribution or the corresponding moments defined in Supplementary Section~S1.

For fine-tuning data auditing, we additionally assume the auditor has black-box query access to the model used to initialize fine-tuning, which serves as a reference model.
This assumption reflects a common scenario in which an organization fine-tunes a publicly available model on private data, while the initial model remains available for querying~\citep{llama}.
Comparing the outputs of the target and reference models provides evidence of changes in model behavior associated with fine-tuning.

\paragraph*{Membership Signals and Evidence}
Training-data auditing uses observable model outputs, such as token-level conditional probabilities, as \emph{membership signals}. 
Auditing methods transform these signals, optionally together with auxiliary information, into scalar statistics that quantify \emph{membership evidence} for a queried sequence. 
For example, perplexity (PPL) summarizes how well the model predicts the sequence, whereas Min-$K\%$ focuses on its least predictable tokens. Smaller PPL values and larger Min-$K\%$ values indicate stronger membership evidence. Other methods incorporate reference-model outputs or textual characteristics to adjust these statistics. 
These methods provide different aspects of the model’s behavior, motivating us to combine their statistics into chunk-level membership evidence for subsequent document-level auditing. 
The individual evidence statistics are detailed in Supplementary Section~S1.

\subsection{Datasets}\label{sec:problem_dataset}

LLM training commonly involves large-scale pre-training followed by fine-tuning on selected data, and our framework audits data use in either stage.

We study BookTection~\citep{duarte2024decop} and Pile--arXiv~\citep{gao2020pile} for pre-training data auditing, and Cosmopedia~\citep{cosmopedia} and HealthCareMagic~\citep[HCM]{chatdoctor} for fine-tuning data auditing. 
Membership is defined with respect to the training stage under audit.

\textbf{BookTection} comprises excerpts from $165$ books, with roughly $34$ original passages per book. We use the benchmark’s Medium setting, which provides passages with a nominal length of $128$ tokens. Each book serves as a document-level inferential unit, with its passages treated as chunks.
Following the benchmark’s publication-date labeling rule, we treat $105$ classic books published in or before $2021$ as candidate members and $60$ books published in $2023$ as non-members. Books published in $2022$ are excluded because their membership status is ambiguous under this rule.
This benchmark allows us to evaluate how aggregating passage-level membership evidence supports individual book auditing.

\textbf{Pile--arXiv} consists of full-text scientific papers from the arXiv component of The Pile~\citep{gao2020pile}.
The public release used here contains $79{,}046$ papers from the first training shard and $4{,}841$ papers from the official held-out validation and test splits. 
Each record contains a paper’s main text, mathematical expressions, references, and source metadata. We treat each paper as a document-level inferential unit.
For Pythia target models trained on the corresponding Pile corpus, papers from the training partition are members, while those from the held-out partitions are non-members.

\textbf{Cosmopedia} provides synthetic English text, including educational explanations, textbook-style passages, and other expository content. We use its \texttt{web\_samples\_v2} subset, which contains $10{,}345{,}867$ records generated from web-page excerpts. 
Each record includes generated text, a generation prompt, and metadata describing the source, format, intended audience, and token length.
We use the generated text as the document-level input and fine-tune Pythia-2.8B on selected documents using the next-token prediction objective.
Documents used in this fine-tuning run are members, while withheld documents are non-members.

\textbf{HealthCareMagic (HCM)} contains $112{,}165$ health-care question--answer records, each comprising an instruction, a patient question, and an answer.
We treat each complete record as a document-level inferential unit.
We adapt Llama-3.2-3B-Instruct to the medical domain through supervised fine-tuning, using the instruction and patient question as conditioning context and the answer as the supervised response.
Records used for gradient updates are members, while withheld records are non-members. %

\subsection{Problem Formulation}\label{sec:problem_formulation}

Let $\mF_{\theta_{\mathrm{target}}}$ denote the target LLM with parameters $\theta_{\mathrm{target}}$ and trained on an undisclosed corpus $\mD_{\mathrm{train}}$.
For a target collection $\mathcal D_{\mathrm{target}}=\{d_i\}_{i=1}^{m}$ of $m$ documents, let $y_i:=\mathbb I\{d_i\in\mD_{\mathrm{train}}\}$ denote the unknown membership label of document $d_i$.
We formulate \emph{training-data auditing} through the following three objectives.

\paragraph*{Problem 1: Document-level individual auditing}
For each target document $d_i$, the auditor seeks to determine whether it was used to train $\mF_{\theta_{\mathrm{target}}}$. Taking the document as the basic inferential unit, we formulate this objective for document $i\in \mD_{target}$ as
\begin{equation}
\label{eq:problem1}
H_{0,i}:\; 
d_i\notin\mD_{\mathrm{train}}
\qquad \text{versus} \qquad
H_{1,i}:\; d_i\in\mD_{\mathrm{train}}.
\end{equation}
Non-membership serves as the null hypothesis, so that a document is flagged as a suspected member only when there is sufficient evidence to reject $H_{0,i}$.

\paragraph*{Problem 2: Collection-level global auditing}
At the collection level, the auditor examines whether the distribution of membership evidence differs between the target collection and known non-members.
Let $\mathcal R^{\mathrm{cal}}=\{d_\ell^{\mathrm{cal}}\}_{\ell=1}^{n_{\mathrm{cal}}}$ denote a calibration collection of $n_{\mathrm{cal}}$ documents drawn from the same underlying distribution as the target documents but confirmed absent from $\mD_{\mathrm{train}}$.
Let $T(\cdot)$ denote the document-level membership statistic, and let $\PP_{\mathrm{target}}$ and $\PP_{\mathrm{cal}}$ denote the distributions of this statistic under the target and calibration populations, respectively.
We formulate global auditing as the two-sample hypothesis test
\begin{equation}
\label{eq:problem2}
H_0^{\mathrm{glob}}:\;
\PP_{\mathrm{target}} = \PP_{\mathrm{cal}}
\qquad \text{versus} \qquad
H_1^{\mathrm{glob}}:\;
\PP_{\mathrm{target}} \neq \PP_{\mathrm{cal}}.
\end{equation}
Rejection indicates a collection-level departure from the evidence distribution of known non-members.
\begin{remark}
An alternative approach combines the individual tests in~\eqnref{eq:problem1}, for example through the Bonferroni rule $\min_i p_i \le \alpha/m$, where $p_i$ is the document-level $p$-value and $\alpha$ is the significance level~\citep{bland1995multiple}.
This rule flags a collection when at least one document provides sufficiently strong evidence after multiplicity adjustment.
As the evidence is usually weak and distributed across several documents, however, no individual $p$-value may reach this threshold even though the collection exhibits a distributional shift. The two-sample formulation aims to capture such collective evidence.
\end{remark}

\paragraph*{Problem 3: Suspicious-subset identification}
A global decision summarizes evidence across a collection but does not identify which documents warrant further investigation.
Suspicious-subset identification addresses this need by returning a data-dependent discovery set $\widehat{\mathcal S}\subseteq \{1,\dots,m\}$ where $m = |\mD_{target}|$, containing the indices of documents flagged as suspected members.
For the prespecified family of all $m$ document-level hypotheses, we seek to control the expected proportion of non-members among the selected documents at a nominal false discovery rate (FDR) level $q\in(0,1)$:
\begin{equation}
\label{eq:problem3}
\operatorname{FDR}(\widehat{\mS}) :=
\mathbb E\left[
\frac{|\widehat{\mS}\cap\mathcal I_0|}{|\widehat{\mS}|\vee1}
\right]
\le q,
\end{equation}
where 
$\mathcal I_0:=\{i:d_i\notin \mathcal D_{\mathrm{train}}\}$
 and 
$\mathcal I_1:=\{i:d_i\in \mathcal D_{\mathrm{train}}\}$
denote the index sets of non-member and member target documents, respectively.
The expectation is taken over the complete reporting procedure under repeated sampling.

%% file: 4_method.tex
\section{The Statistical Multi-Level Auditing Framework}\label{sec:method}

\begin{figure}[!htbp]
    \centering
    \includegraphics[width=0.9\linewidth,trim={23bp 24bp 23bp 24bp},clip]{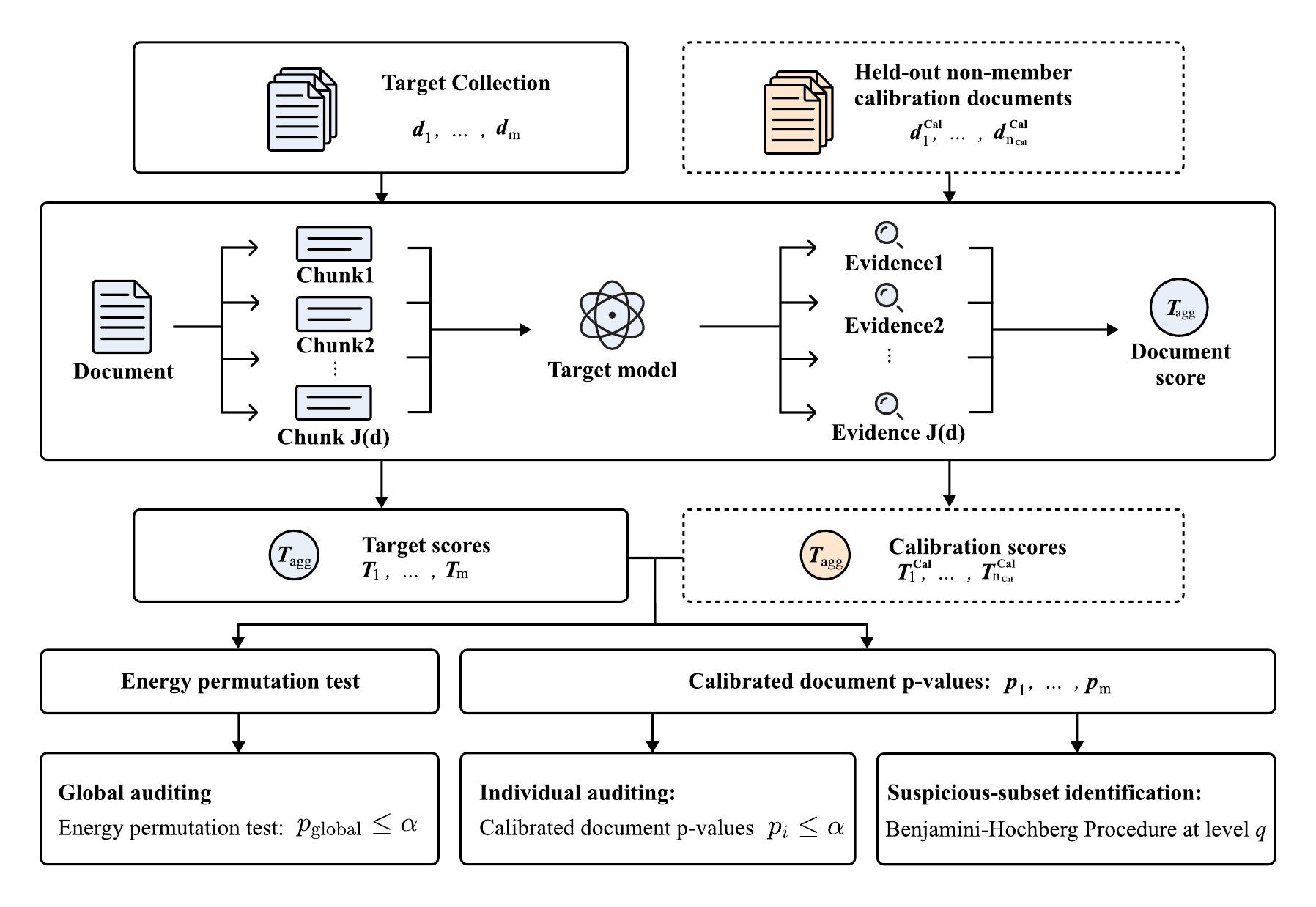}
    \caption{Multi-level statistical framework for LLM training-data auditing. Target documents and held-out non-member calibration documents follow the same scoring pipeline: each document is split into chunks, membership evidence is extracted from the target model for each chunk, and the evidence is aggregated into a document score. These scores support three auditing tasks. An Energy permutation test compares the target and calibration score distributions for global auditing. Empirical calibration yields document-level $p$-values for individual auditing, and applying the Benjamini--Hochberg procedure to these $p$-values identifies a suspicious subset.}
    \label{fig:teaser}
\end{figure}

In the proposed framework, documents serve as inferential units, while \emph{chunks} serve as operational units for extracting membership evidence~\citep{wbc}.
The overall pipeline is summarized in \figref{fig:teaser}. 
We first combine membership evidence for each chunk and aggregate it into a document-level score. Empirical calibration of this score yields a $p$-value for individual membership testing. We then compare target and non-member document-score distributions through an Energy permutation test for global auditing, and apply BH to the document-level $p$-values for suspicious-subset identification.

\subsection{Constructing Chunk-Level Membership Evidence}\label{sec:method_chunk_evidence}

For each target document $d_i\in\mathcal D_{\mathrm{target}}$, we apply a fixed chunking rule to obtain $\operatorname{Chunk}(d_i)=\bigl\{c_j(d_i)\bigr\}_{j=1}^{J(d_i)}$, where $J(d_i)$ is the number of chunks and $c_j(d_i)$ is the $j$-th chunk.
For each chunk $c$, we compute $d$ membership-evidence statistics, $(e_1(c),\ldots,e_d(c))$, from the available model outputs and auxiliary information.
To define these statistics, write a chunk as a token sequence $c=(w_1,\ldots,w_L)$. Examples include the following:
\begin{itemize}
    \item \emph{Perplexity score,} defined as the exponential of the average per-token NLL,
    \begin{equation}
    e_{\mathrm{PPL}}(c) := \exp\Bigl\{-\frac{1}{L}\sum_{t=1}^{L}\log p_{\theta_{\mathrm{target}}}\bigl(w_t\mid\mathbf w_{<t}\bigr)\Bigr\};
    \notag
\end{equation}
    with smaller values indicating stronger membership evidence.
    \item \emph{Min-$K\%$ score~\citep{mink},} defined as the average log-probability over the least predictable tokens,
    \begin{equation}
    e_{\mathrm{Min}\text{-}K,\rho}(c) := \frac{1}{|\mathcal T_\rho(c)|}\sum_{t\in\mathcal T_\rho(c)}\log p_{\theta_{\mathrm{target}}}\bigl(w_t\mid\mathbf w_{<t}\bigr),
    \notag
\end{equation}
    where $\mathcal T_\rho(c)$ contains the indices of the $\rho\%$ tokens with the smallest log-probabilities. Larger values indicate stronger membership evidence.
\end{itemize}
Additional evidence statistics are defined in Supplementary Section~S1.
These statistics differ in scale and evidence direction, and their relative usefulness may vary across auditing settings.
We therefore learn a combination that converts them into a scalar chunk-level membership evidence $e_{\mathrm{com}}(c)$.

We fit this combination using an auxiliary \emph{fitting collection} $\mathcal R^{\mathrm{fit}}=\{(d_k^{\mathrm{fit}},y_k^{\mathrm{fit}})\}_k$, comprising a small set of documents with membership labels known to the auditor.
Such labels may be available through interaction with the model developer to establish which documents were used or withheld during training.
We apply the same chunking rule to each fitting document and assign its label $y_k^{\mathrm{fit}}$ to every resulting chunk $c_j(d_k^{\mathrm{fit}})$.

Let $y(c)$ denote the label assigned to chunk $c$, where $y(c)=1$ indicates membership and $y(c)=0$ non-membership.
 We model its membership probability using logistic regression,
\begin{equation}
\label{eq:chunk_logistic}
\Pr\{y(c)=1 \mid e_1(c),\ldots,e_d(c)\} = \sigma\!\left(\beta_0+\sum_{r=1}^{d}\beta_r e_r(c)\right),
\end{equation}
where $\sigma(z)=1/(1+e^{-z})$ is the sigmoid function, $\beta_0$ is the intercept, and $\beta_r$ is the coefficient associated with the $r$-th evidence statistic.
The coefficients are estimated using the pooled chunk–label pairs $(c_j(d_k^{\mathrm{fit}}),y_k^{\mathrm{fit}})$ from the fitting collection. Their signs and magnitudes determine how the evidence statistics contribute to the combined score.

After fitting, we freeze the evidence map and use its logit as the combined chunk-level membership evidence:
\begin{equation}
\label{eq:chunk_evidence}
\begin{aligned}
e_{\mathrm{com}}(c) = \log \frac{\widehat{\Pr}\{y(c)=1 \mid e_1(c),\ldots,e_d(c)\}} {\widehat{\Pr}\{y(c)=0 \mid e_1(c),\ldots,e_d(c)\}} = \widehat{\beta}_0+\sum_{r=1}^{d}\widehat{\beta}_r e_r(c).
\end{aligned}
\end{equation}
Larger values indicate stronger membership evidence.
The logit of the conditional class probabilities approximates the log-likelihood ratio between the member-labeled and non-member-labeled evidence distributions, up to an additive constant determined by their class proportions.
This interpretation connects the learned evidence to likelihood-ratio-based membership inference~\citep{lira}.

\subsection{Cauchy Combination for Document-Level Individual Auditing}\label{sec:method_doc_level}

Document-level auditing requires aggregating the chunk-level evidence into a single document score. Membership evidence may be concentrated in a few passages, while the remaining chunks contribute weak or uninformative evidence~\citep{wbc}. Simply averaging chunk-level evidence~\citep{duarte2024decop} can dilute these localized signals. Moreover, chunks within a document may be dependent because they share topics and context. We use Cauchy combination~\citep{liu2020cauchy} to construct a document score sensitive to strong local evidence, followed by empirical document-level calibration for individual testing.

We first convert each chunk's combined evidence $e_{\mathrm{com}}(c)$ into an empirical $p$-value, which requires the  distribution of $e_{\mathrm{com}}$ under non-membership.
We introduce a \emph{base collection} $\mathcal R^{\mathrm{base}}$ of known non-member documents and apply the same chunking rule to obtain $n_{\mathrm{base}}$ chunks, $\{c_h^{\mathrm{base}}\}_{h=1}^{n_{\mathrm{base}}}$.
Their evidence values, computed using the frozen evidence map, provide an empirical reference for non-member behavior. 
For each target chunk $c_j(d_i)$, we compute the empirical right-tail $p$-value,
\begin{equation}
\label{eq:chunk_pvalue}
p_{ij}^{\mathrm{ck}}
=
\frac{
1+\sum_{h=1}^{n_{\mathrm{base}}}
\mathbb I\!\left\{
e_{\mathrm{com}}(c_h^{\mathrm{base}})
\ge e_{\mathrm{com}}(c_j(d_i))
\right\}
}{
n_{\mathrm{base}}+1
},
\quad i=1,\ldots,m,\quad j=1,\ldots,J(d_i).
\end{equation}
A smaller $p_{ij}^{\mathrm{ck}}$ indicates that the target chunk exhibits stronger membership evidence relative to the non-member base chunks.

We aggregate the empirical chunk-level $p$-values $\{p_{ij}^{\mathrm{ck}}\}_{j=1}^{J(d_i)}$ computed for the $J(d_i)$ chunks $\{c_j(d_i)\}_{j=1}^{J(d_i)}$ of document $d_i$ into the document score
\begin{equation}
\label{eq:doc_cauchy}
T_{\mathrm{Cauchy}}(d_i) = \frac{1}{J(d_i)}\sum_{j=1}^{J(d_i)}
\cot \bigl(\pi[p_{ij}^{\mathrm{ck}}]_{\varepsilon}\bigr),
\end{equation}
where 
$\cot(\pi p)=\tan\bigl((0.5-p)\pi\bigr)$ is the Cauchy transformation and $[p]_{\varepsilon}=\min\{1-\varepsilon,\max\{\varepsilon,p\}\}$ clips $p$ to $[\varepsilon,1-\varepsilon]$.
The clipping parameter $\varepsilon\in(0,1/2)$ is fixed in advance and shared across all target and calibration documents, ensuring numerical stability.
The Cauchy transformation maps small $p$-values to large positive contributions, values near $0.5$ to contributions near zero, and values above $0.5$ to negative contributions.
Consequently, sufficiently small $p$-values from a few informative chunks can contribute substantially to the document score, allowing localized membership signals to drive document-level decisions.
Larger values of $T_{\mathrm{Cauchy}}(d_i)$ indicate stronger membership evidence.

To obtain an individual auditing decision, we calibrate the document score using the held-out non-member collection $\mathcal R^{\mathrm{cal}}=\{d_\ell^{\mathrm{cal}}\}_{\ell=1}^{n_{\mathrm{cal}}}$.
This collection is held out from fitting and selecting the complete scoring rule.  
Let $T_{\mathrm{agg}}$ denote the fixed document scoring rule, with $T_{\mathrm{agg}}=T_{\mathrm{Cauchy}}$ for the construction above.
Applying the same rule to target and calibration documents gives the empirical right-tail $p$-value
\begin{equation}
\label{eq:doc_pvalue}
p_i =
\frac{
1+\sum_{\ell=1}^{n_{\mathrm{cal}}}
\mathbb I\!\left\{
T_{\mathrm{agg}}(d_\ell^{\mathrm{cal}})
\ge T_{\mathrm{agg}}(d_i)
\right\}
}{
n_{\mathrm{cal}}+1
},
\qquad i=1,\ldots,m.
\end{equation}
At a prespecified significance level $\alpha\in(0,1)$, we reject $H_{0,i}$ when $p_i\leq\alpha$. 
Alternative aggregation rules are discussed and compared in Section~S2 of the  Supplementary Materials.

\begin{remark}
For the Cauchy score, an analytic counterpart of \eqnref{eq:doc_pvalue} is available without any calibration data. If the chunk-level $p$-values were independent and uniformly distributed under the null, and no clipping were applied,
$T_{\mathrm{Cauchy}}(d_i)$ would follow a standard Cauchy distribution under $H_{0,i}$, and the upper-tail probability
\begin{equation}
\label{eq:doc_pvalue_cct}
u(d_i) := 1-F_{\mathrm{Cauchy}}\{T_{\mathrm{Cauchy}}(d_i)\} = \tfrac12-\tfrac1\pi\arctan\{T_{\mathrm{Cauchy}}(d_i)\}
\end{equation}
would be an exact document-level $p$-value~\citep{liu2020cauchy}.
Our chunk-level $p$-values are empirical and dependent, and the Cauchy transformation is clipped. We therefore use the empirical calibration in \eqnref{eq:doc_pvalue} for individual testing and subset identification, whose validity under conditional exchangeability is established in Theorem~\ref{thm:document}. The analytic form \eqnref{eq:doc_pvalue_cct} is used below as a bounded score transformation for the Energy two-sample comparison. 
\end{remark}

\subsection{Energy-Based Two-Sample Testing for Global Auditing}
\label{sec:method_global_test}

Building on the document-level scores above, we now address global auditing through the two-sample testing problem formulated in \eqnref{eq:problem2}.
Specifically, we compare the document-score distributions of $\mathcal D_{\mathrm{target}}$ and the held-out non-member calibration collection $\mathcal R^{\mathrm{cal}}$.
For the Cauchy document statistic in \eqnref{eq:doc_cauchy}, we apply the bounded transformation in \eqnref{eq:doc_pvalue_cct}, writing $u(d):= \tfrac12-\tfrac1\pi\arctan\{T_{\mathrm{Cauchy}}(d_i)\}$.
Its bounded range $(0,1)$ improves the numerical stability of the subsequent distance computations.

Let $F_{\mathrm{target}}$ and $F_{\mathrm{cal}}$ denote the population distributions of $u(d)$ for target and calibration documents, respectively.
Following \eqnref{eq:problem2}, we test $H_0^{\mathrm{glob}}:F_{\mathrm{target}}=F_{\mathrm{cal}}$ against $H_1^{\mathrm{glob}}:F_{\mathrm{target}}\neq F_{\mathrm{cal}}$.
We measure the distributional difference using the energy discrepancy~\citep{szekely2013energy},
\begin{equation}
\label{eq:energy_distance}
\mathcal E(F_{\mathrm{target}},F_{\mathrm{cal}})
=
2\mathbb E|U-V|
-\mathbb E|U-U'|
-\mathbb E|V-V'|,
\end{equation}
where $U,U'$ are independent draws from $F_{\mathrm{target}}$, $V,V'$ are independent draws from $F_{\mathrm{cal}}$, and the two pairs are independent.
The discrepancy is zero if and only if the two distributions coincide.
We compute the sample energy statistic $\mathcal E_{\mathrm{obs}}$ from $\{u(d_i)\}_{i=1}^{m}$ and $\{u(d_\ell^{\mathrm{cal}})\}_{\ell=1}^{n_{\mathrm{cal}}}$, with its implementation specified in Supplementary Section~S4.7.

We obtain the collection-level $p$-value by a permutation procedure~\citep{phipson2010permutation,hemerik2018exact}: drawing $B$ uniform permutations of the document-level group assignments, each producing a permuted statistic $\mathcal E_b$ with $b=1,\dots, B$, we compute
\begin{equation}
\label{eq:energy_global_pvalue}
p_{\mathrm{global}}
=
\frac{1+\sum_{b=1}^{B}\mathbb I\{\mathcal E_b\geq \mathcal E_{\mathrm{obs}}\}}{B+1}.
\end{equation}
At a prespecified significance level $\alpha$, we reject $H_0^{\mathrm{glob}}$ when $p_{\mathrm{global}}\leq\alpha$.
Rejection indicates that the target collection differs from the non-member calibration population in the distribution of document scores.

\begin{remark}
Existing mean-based two-sample comparisons, such as the $t$-test used for dataset inference~\citep{maini2024datasetllm}, focus on differences in location and need not detect changes in dispersion or shape. The energy discrepancy therefore provides a broader basis for global auditing, since it responds to any of these departures rather than to location shifts alone.\end{remark}%

\subsection{BH Procedure for Suspicious-Subset Identification}
\label{sec:method_fdr_control}

Suspicious-subset identification seeks to determine which target documents warrant further investigation while controlling false discoveries.
Following Problem 3, we formulate this task as multiple testing of the complete prespecified family of $m$ document-level hypotheses.
We apply the Benjamini--Hochberg (BH) procedure~\citep{benjamini1995controlling}, with the false discovery rate measuring the expected proportion of non-members among the selected documents.

Let $p_{(1)}\leq\cdots\leq p_{(m)}$ denote the ordered
document-level $p$-values, with corresponding document indices $i_{(1)},\ldots,i_{(m)}$.
At the prespecified FDR level $q$, BH finds the largest rank satisfying
\begin{equation}
\label{eq:by_rule}
R_{\mathrm{BH}} =
\max\left\{r\in\{1,\ldots,m\}:p_{(r)}\le\frac{r}{m}q\right\}.
\end{equation}
If no rank satisfies the inequality, we set $R_{\mathrm{BH}}=0$.
The resulting discovery set is $$\widehat{\mS}_{\mathrm{BH}}
=\{i_{(1)},\ldots,i_{(R_{\mathrm{BH}})}\},$$ with $\widehat{\mS}_{\mathrm{BH}}=\varnothing$
when $R_{\mathrm{BH}}=0$.

The empirical $p$-values are dependent because they share the same calibration collection. Classical BH theory provides FDR control under positive regression dependence on the subset of true null hypotheses~\citep{benjamini2001control}. For the empirical calibration in \eqnref{eq:doc_pvalue}, Theorem~\ref{thm:by} establishes finite-sample FDR control under Assumption~\ref{ass:document}(i): conditional on the fixed scoring procedure and the scores of member target documents, the calibration and non-member target scores are jointly exchangeable. Under this condition, applying BH to all $m$ hypotheses controls the expected proportion of non-members in $\widehat{\mathcal S}_{\mathrm{BH}}$ at level $q$.

%% file: 5_theorey.tex
\section{Theoretical Guarantees}\label{sec:theoretical_guarantees}

We study the statistical properties of the proposed auditing framework. 
We first characterize how localized chunk-level evidence contributes to the document score. 
Under conditional exchangeability, we then establish the validity of empirical document-level $p$-values, type-I error control for Energy-based global auditing, and FDR control for BH subset identification. 
Finally, we quantify how departures from exchangeability affect these guarantees.
Proofs are provided in Supplementary Section~S3.

Recall that the target collection $\mathcal D_{\mathrm{target}}=\{d_i\}_{i=1}^{m}$ contains $m$ documents, each document $d_i$ comprising $J(d_i)$ chunks. The base collection $\mathcal R^{\mathrm{base}}$ supplies $n_{\mathrm{base}}$ non-member chunks for chunk-level calibration.
The calibration collection $\mathcal R^{\mathrm{cal}}=\{d_\ell^{\mathrm{cal}}\}_{\ell=1}^{n_{\mathrm{cal}}}$ contains $n_{\mathrm{cal}}$ non-member documents used for document-level calibration and the comparison sample in global auditing.

\subsection{Localized Chunk-Level Evidence}
\label{sec:theory_chunk}

Membership evidence may be concentrated in a small subset of a document’s chunks. 
The Cauchy combination allows these chunks to influence the document score even when other chunks contribute weak evidence~\citep{liu2020cauchy}.
The following proposition quantifies this contribution.

\begin{proposition}[Sensitivity to localized evidence]
\label{prop:localized}
Let $S_i\subseteq\{1,\ldots,J(d_i)\}$ be a nonempty subset of chunk indices. Suppose that $p^{\mathrm{ck}}_{ij}\leq u$ for all $j\in S_i$. 
If the remaining chunks satisfy $\sum_{j\notin S_i}\cot\!\bigl(\pi[p^{\mathrm{ck}}_{ij}]_{\varepsilon}\bigr)\geq-\bigl(J(d_i)-|S_i|\bigr)b$, where $\varepsilon\leq u<1/2$ and
  $b\geq0$, then
\begin{equation}
\label{eq:localized_bound}
T_{\mathrm{Cauchy}}(d_i)
\geq
\frac{|S_i|}{J(d_i)}\cot(\pi u)
-
\frac{J(d_i)-|S_i|}{J(d_i)}b.
\end{equation}
The first term bounds the contribution from the informative chunks in $S_i$, while the second bounds the negative offset from the remaining chunks.
\end{proposition}

Proposition~\ref{prop:localized} characterizes the contribution of localized chunk-level evidence to the Cauchy document score. 
For small $u\geq\varepsilon$, the positive term is approximately $|S_i|/\{\pi J(d_i)u\}$. 
By comparison, averaging $1-p^{\mathrm{ck}}_{ij}$ across chunks limits the contribution from $S_i$ to at most $|S_i|/J(d_i)$.
Cauchy aggregation therefore amplifies sufficiently small chunk-level  $p$-values. This amplification allows a small subset of chunks to yield a positive document score when its contribution outweighs the negative offset from the remaining chunks. 
The normalization by $J(d_i)$ remains: the contribution depends on both the strength of the localized evidence and the proportion of chunks carrying it.

\subsection{Error Control under Conditional Exchangeability}\label{sec:theory_document}\label{sec:theory_collection}

Let $T_{\mathrm{agg}}$ be a measurable real-valued document scoring rule, such as the Cauchy-combined score in \eqnref{eq:doc_cauchy}.
Let $\mathcal G$ denote the sigma-field generated by the target model, the fitting collection, the fitted evidence map, the base collection, the chunking rule, and the aggregation rule.
Conditional on $\mathcal G$, the complete scoring procedure is fixed and applied identically to all documents.
The remaining randomness comes from the target and calibration collections.

The error guarantees rely on permutation symmetry among the relevant document scores. We state the conditions separately for document-level calibration and subset identification, and for global auditing.

\begin{assumption}[Conditional exchangeability]\label{ass:document}
\begin{enumerate}
\item[(i)] Conditional on $\mathcal G$ and the scores of all member target documents, the scores of all calibration documents and all non-member target documents are jointly exchangeable.
\item[(ii)] Under the global null $H_0^{\mathrm{glob}}$, the scores of all target and calibration documents are jointly exchangeable conditional on $\mathcal G$.
\end{enumerate}
\end{assumption}

Conditional on $\mathcal G$, exchanging the positions of calibration and non-member target scores leaves their joint distribution unchanged.
Condition (i) requires this symmetry to hold even after conditioning on the member target scores. This additional requirement supports BH error control because its rejection threshold depends on both member and non-member scores.
The symmetry allows calibration scores to provide a reference for non-member target scores, even when these scores are dependent.
Condition (ii) imposes the corresponding symmetry on all target and calibration scores under the global null, providing the basis for the reassignment of document-level group labels in Energy permutation testing.

The document-level $p$-values share the same calibration collection and are therefore generally dependent~\citep{bates2023outliers}.
Our error guarantees accommodate this dependence under the conditional exchangeability assumptions on the document scores.

The following three theorems establish document-level $p$-value validity, global type-I error control, and FDR control for suspicious-subset identification, corresponding to Problems 1, 2, and 3, respectively.

\Needspace{7\baselineskip}
\begin{theorem}[Validity of document-level $p$-values]\label{thm:document}
Under Assumption~\ref{ass:document}(i), for every true null $H_{0,i}$ and every $\xi\in[0,1]$, the empirical document-level $p$-value in \eqnref{eq:doc_pvalue} satisfies
\begin{equation}\label{eq:document_valid}
 \Pr_{H_{0,i}}(p_i\leq\xi\mid\mathcal G)
 \leq\frac{\lfloor(n_{\mathrm{cal}}+1)\xi\rfloor}{n_{\mathrm{cal}}+1}
 \leq\xi.
\end{equation}
If the target document score and all calibration document scores are distinct almost surely, equality holds in the first inequality. Consequently, rejecting $H_{0,i}$ when $p_i\leq\alpha$ controls its conditional type-I error at any level $\alpha\in(0,1)$.
\end{theorem}

Theorem~\ref{thm:document} establishes validity after any fixed measurable document scoring rule satisfying the exchangeability condition. Under this symmetry, a non-member target score is compared with calibration scores on the same basis; when scores are distinct, its rank is uniformly distributed over the $n_{\mathrm{cal}}+1$ possible positions. 
The floor term in \eqnref{eq:document_valid} reflects the discrete resolution of empirical calibration. For example, when $n_{\mathrm{cal}}=99$, the possible $p$-values are $0.01,0.02,\ldots,1$. Testing at $\alpha=0.025$ is therefore equivalent to rejecting when $p_i\leq0.02$, giving a conditional type-I error of at most $0.02$. This bound is attained when scores are almost surely distinct.
Counting calibration scores equal to the target score in the right tail preserves validity and can make the test more conservative.

For global auditing, the same fixed transformation is applied to target and calibration scores before computing the Energy statistic.
This transformation preserves the conditional exchangeability in Assumption~\ref{ass:document}(ii), yielding the following guarantee.

\Needspace{7\baselineskip}
\begin{theorem}[Type-I error control for global auditing]\label{thm:global}
Under Assumption~\ref{ass:document}(ii), let $p_{\mathrm{global}}$ be the Energy permutation $p$-value computed using $B$ independent uniform permutations, drawn independently of the data and preserving the target and calibration group sizes. 
For every $\alpha\in(0,1)$, define $\mathcal E_\alpha=\{p_{\mathrm{global}}\leq\alpha\}$. Then
\begin{equation}\label{eq:global_exact}
 \Pr_{H_0^{\mathrm{glob}}}(\mathcal E_\alpha\mid\mathcal G)
 \leq\frac{\lfloor(B+1)\alpha\rfloor}{B+1}
 \leq\alpha.
\end{equation}
\end{theorem}

Theorem~\ref{thm:global} provides type-I error control for testing equality of the target and calibration score distributions. 

Suspicious-subset identification requires controlling false discoveries across the complete target family. The next result establishes this guarantee for BH applied to the empirically calibrated document-level $p$-values.

\begin{theorem}[FDR control for suspicious-subset identification]\label{thm:by}
Under Assumption~\ref{ass:document}(i), applying BH to all $m$ empirical document-level $p$-values at level $q\in(0,1)$ gives
\begin{equation}\label{eq:by_exact}
\operatorname{FDR}(\widehat{\mS}_{\mathrm{BH}}\mid\mathcal G)=\mathbb E\!\left[\frac{|\widehat{\mS}_{\mathrm{BH}}\cap\mathcal I_0|}{|\widehat{\mS}_{\mathrm{BH}}|\vee1}\;\middle|\;\mathcal G\right]\leq\frac{|\mathcal I_0|}{m}q\leq q,
\end{equation}
where $\mathcal I_0$ is the set of non-member target indices.
The bound holds when calibration scores equal to a target score are included in its empirical right-tail count.
\end{theorem}

Theorem~\ref{thm:by} extends individual error control to simultaneous selection across the target collection. Under Assumption~\ref{ass:document}(i), it accounts for the shared calibration collection and the data-dependent BH threshold, bounding the expected proportion of non-members among the selected documents.

\subsection{Robustness to Approximate Exchangeability}\label{sec:theory_robust}

The preceding error guarantees rely on conditional exchangeability.
We now quantify how departures from this condition affect these guarantees.
We measure these departures by the total variation distance between the actual joint score distribution and a reference distribution satisfying the corresponding exchangeability condition.

Conditional on $\mathcal G$, let $\mathsf J$ denote the joint distribution of all target and calibration document scores, and let $\mathsf J_\star$ be an ideal counterpart.
Membership labels are held fixed under both distributions.
For document-level testing and BH subset identification, $\mathsf J_\star$ satisfies Assumption~\ref{ass:document}(i); for global testing under $H_0^{\mathrm{glob}}$, it satisfies Assumption~\ref{ass:document}(ii). For some $\delta\in[0,1]$, suppose that, almost surely,
\begin{equation}
d_{\mathrm{TV}}(\mathsf J,\mathsf J_\star) =
\sup_A\left|\mathsf J(A)-\mathsf J_\star(A)\right|
\leq\delta,
\notag
\end{equation}
where the supremum is over measurable events. Thus, the actual and ideal conditional joint distributions assign probabilities differing by at most $\delta$ to any event.

\Needspace{7\baselineskip}     
\begin{theorem}[Error control under $\delta$-approximate exchangeability]\label{thm:robust}
Under the joint total variation bound and the corresponding exchangeability conditions for $\mathsf J_\star$, the following guarantees hold.
For every $i\in\mathcal I_0$ and every $\xi\in[0,1]$,
\begin{equation}\label{eq:document_approx}
\Pr(p_i\leq\xi\mid\mathcal G)
\leq
\min\left\{
1,\frac{\lfloor(n_{\mathrm{cal}}+1)\xi\rfloor}{n_{\mathrm{cal}}+1}+\delta
\right\}.
\end{equation}
For the BH discovery set at level $q\in(0,1)$,
\begin{equation}\label{eq:by_approx}
\operatorname{FDR}(\widehat{\mS}_{\mathrm{BH}}\mid\mathcal G)
\leq
\min\left\{
1,\frac{|\mathcal I_0|}{m}q+\delta
\right\}.
\end{equation}
Under $H_0^{\mathrm{glob}}$, the Energy permutation test in Theorem~\ref{thm:global} satisfies
\begin{equation}\label{eq:global_approx}
\Pr_{H_0^{\mathrm{glob}}}(\mathcal E_\alpha\mid\mathcal G)\leq\min\left\{1,\frac{\lfloor(B+1)\alpha\rfloor}{B+1}+\delta\right\}\leq\min\{1,\alpha+\delta\}.
\end{equation}
\end{theorem}

Theorem~\ref{thm:robust} identifies $\delta$ as an additional error cost of departure from exchangeability.
A discrepancy of at most $\delta$ between the conditional joint score distributions increases each of the document-level type-I error, global type-I error, and BH FDR bounds by at most $\delta$.
The guarantees therefore remain close to their exact-exchangeability bounds when this joint discrepancy is small.
This result motivates the use of calibration populations representative of non-member targets and sampling schemes compatible with the required permutation symmetry.
Supplementary Section~S7 examines empirical calibration robustness on HCM by varying the number and answer-length composition of non-member calibration documents.

\paragraph*{Summary.}
The analysis characterizes how localized chunk-level evidence contributes to document scoring and establishes finite-sample error guarantees for the three auditing objectives. 
Under the stated conditional exchangeability conditions, empirical calibration supports valid individual testing, Energy permutation testing controls global type-I error, and BH controls FDR for suspicious-subset identification. 
The robustness bounds quantify the additional error cost when the joint score distribution departs from these conditions.

%% file: 6_exp.tex
\section{Experiment}\label{sec:exp}

We evaluate the training-data-auditing framework through four case studies covering two training stages:
\begin{itemize}
    \item \textbf{Pretraining.} We study copyrighted books from BookTection (Supplementary Section~S5) and scientific papers from Pile–arXiv (\secref{sec:exp_arxiv}) to examine how combining and aggregating passage-level evidence supports long-document auditing.
    \item \textbf{Fine-tuning.} We study synthetic text from Cosmopedia (\secref{sec:exp_cosmopedia}) and medical question–answer records from HealthCareMagic (HCM; \secref{sec:exp_hcm}) to evaluate membership evidence associated with model adaptation and its use for individual and collection-level decisions.
\end{itemize}

The three case studies in the main text each address all three auditing objectives: document-level individual auditing (Problem 1), collection-level global auditing (Problem 2), and suspicious-subset identification (Problem 3). The BookTection study in the Supplementary Material focuses on Problem 1.

For individual auditing, we report the area under the receiver operating characteristic curve (AUC) and the true positive rate (TPR) at specified false positive rates (FPRs), assessing overall discrimination and identification under restricted false-positive budgets.
For global auditing, we report the rejection rate at $\alpha=0.05$, defined as the proportion of sampled target collections for which the global null hypothesis is rejected.
For suspicious-subset identification, we report empirical power and empirical false discovery rate (FDR) at $q=0.10$, measuring the average proportion of true members identified and the average proportion of non-members among selected documents, respectively.

Collection-level experiments use $1{,}000$ sampled target collections per configuration, with collection sizes $m\in\{30,50,100\}$ and member ratios from $10\%$ to $100\%$.
We additionally evaluate entirely non-member collections to assess false positives in global auditing.
Detailed experimental settings and evaluation protocols are provided in Supplementary Section~S4.

\subsection{Detecting Scientific Papers in Pretraining Data}
\label{sec:exp_arxiv}

We evaluate the methods on the Pile–arXiv dataset using Pythia-6.9B~\citep{biderman2023pythia}. Each document is represented by eight $1{,}024$-token chunks. 
For individual auditing, we compare with perplexity (PPL)~\citep{loss}, Min-$K\%$~\citep{mink}, and Mean-evidence to examine evidence combination and aggregation in scientific text. 
We also include Mann--Whitney U, a nonparametric rank test to aggregate chunk-level scores~\citep{mia_scaling}.
Cauchy aggregation outperforms PPL and Min-$K\%$ across all three reported metrics and achieves higher TPRs than Mann--Whitney U at both FPR levels (Table~\ref{tab:arxiv_auc}).
Compared with Mean-evidence, it yields a slightly lower AUC ($75.54\%$ versus $76.04\%$) but improves TPR at $1\%$ FPR from $7.40\%$ to $15.60\%$ and at $5\%$ FPR from $19.80\%$ to $23.00\%$.
The improvement at $1\%$ FPR is relevant to practical auditing, where identifying members under a stringent false-positive budget is a central concern.
Additional comparisons of document-scoring rules are provided in the BookTection study in Supplementary Section~S5.

\input{arxiv_results.tex}

For global auditing, Figure~\ref{fig:arxiv_global} shows that the energy-based test generally achieves higher rejection rates as the member ratio and collection size increase.
These results indicate that a collection can exhibit a detectable distributional departure when evidence is accumulated across multiple papers, even when individual-paper discrimination remains moderate.
Moreover, the dependence on collection size is consistent with this role of global auditing: increasing the number of documents strengthens the comparison between target and non-member score distributions.

\begin{figure}[!htbp]
    \setlength{\parskip}{0pt}
    \captionsetup{skip=4pt}
    \captionsetup[subfigure]{skip=2pt}
    \renewcommand{\thesubfigure}{\alph{subfigure}}
    \centering
    \begin{subfigure}[t]{0.49\linewidth}
        \centering
        \includegraphics[width=\linewidth]{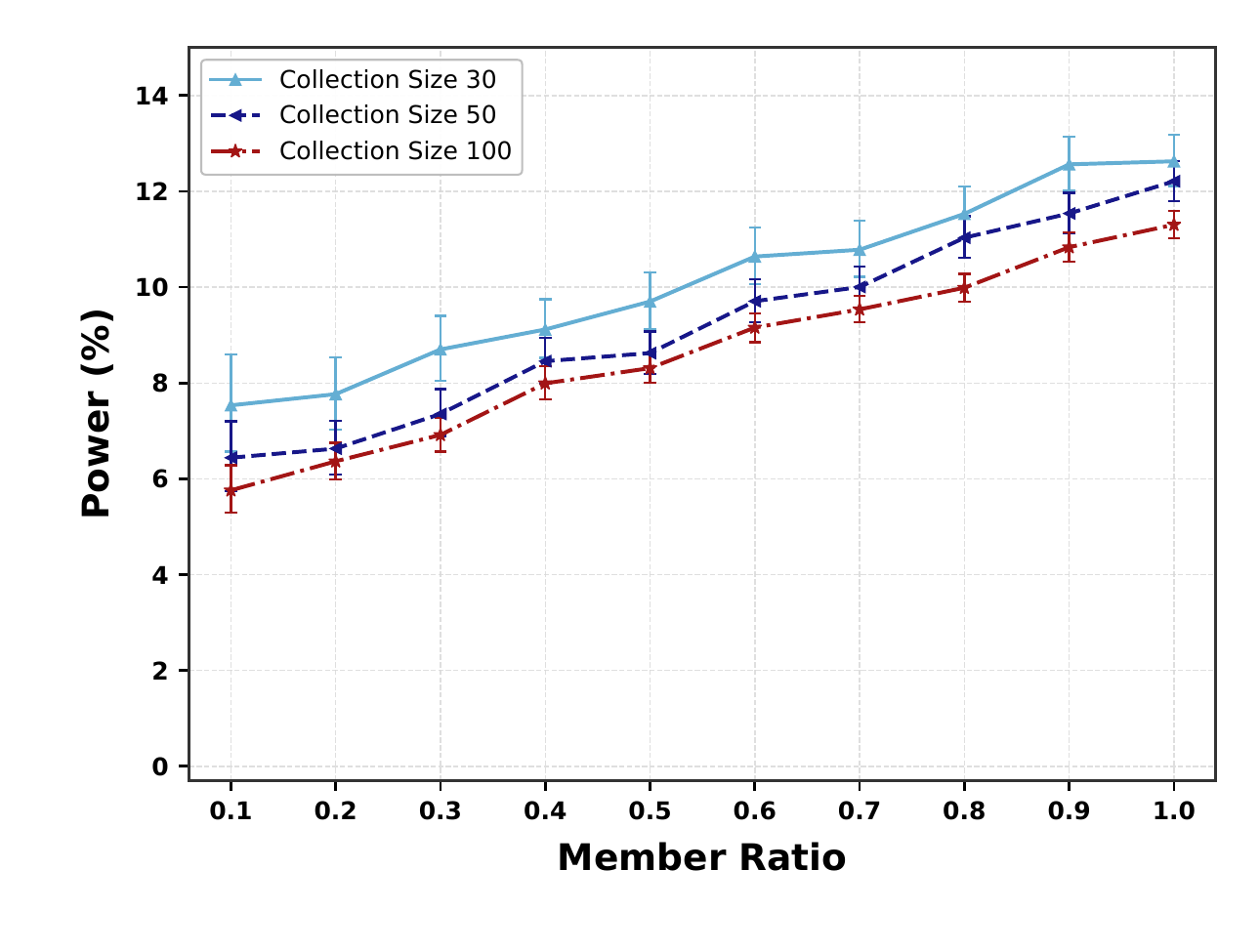}
        \caption{Power}
        \label{fig:arxiv_subset_recall}
    \end{subfigure}
    \hfill
    \begin{subfigure}[t]{0.49\linewidth}
        \centering
        \includegraphics[width=\linewidth]{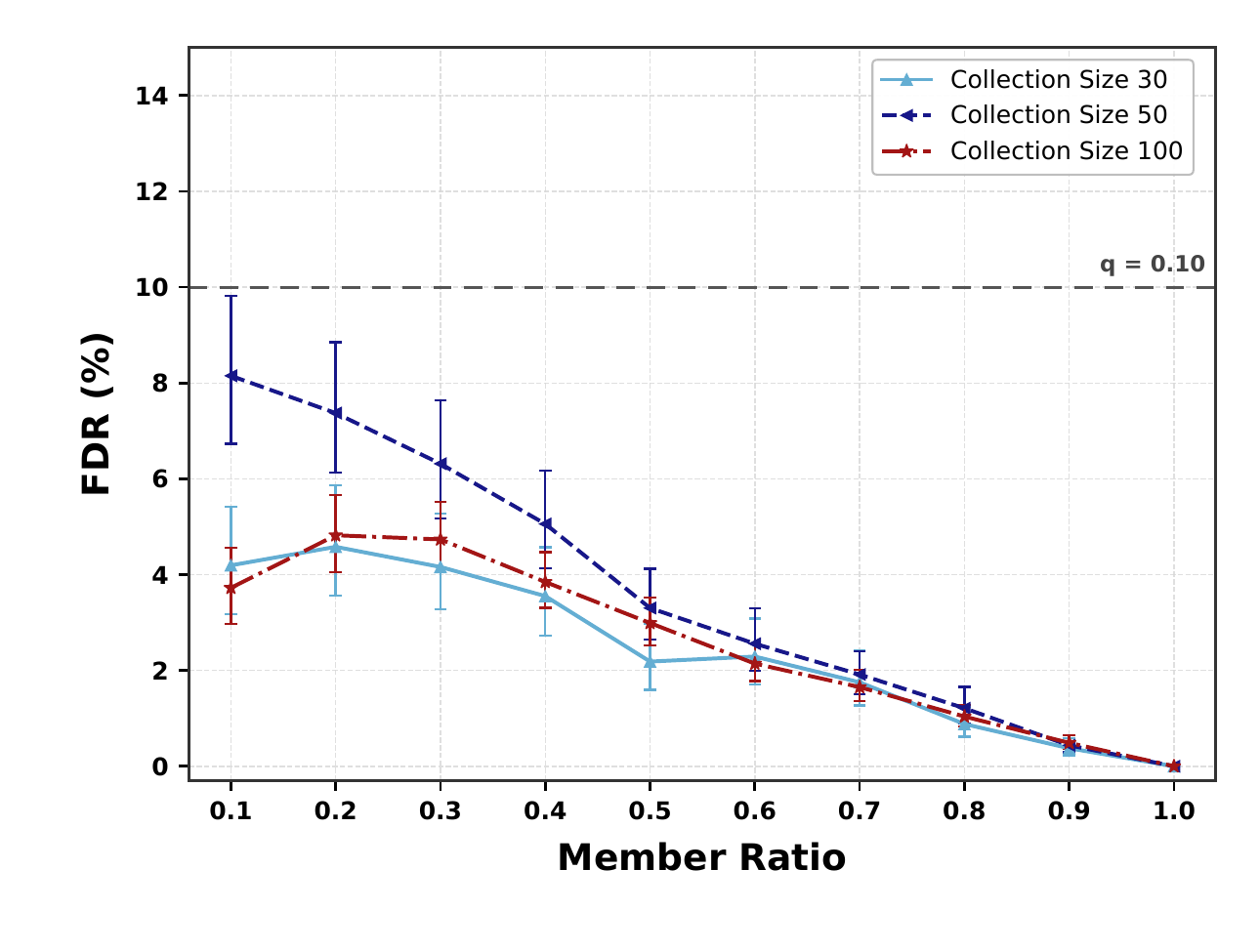}
        \caption{FDR}
        \label{fig:arxiv_subset_fdr}
    \end{subfigure}
    \caption{BH subset identification on Pile–arXiv. Power and empirical FDR are the respective averages of the within-collection true positive fraction and false discovery proportion over $1{,}000$ collections per configuration. Error bars show pointwise $95\%$ BCa bootstrap confidence intervals. The dashed line marks the target FDR level.}
    \label{fig:arxiv_subset}
\end{figure}

For suspicious-subset identification, we apply the BH procedure at $q=0.10$.
Figure~\ref{fig:arxiv_subset} shows that empirical FDR estimates remain below the nominal $10\%$ level across all evaluated configurations.
Power increases with the member ratio and is higher for smaller collections, but remains approximately $11$--$13\%$ even when all target documents are members.
The distinction between global detection and subset identification is particularly clear at a $50\%$ member ratio: the Energy rejection rate increases from approximately $75\%$ to $100\%$ as collection size increases from $30$ to $100$, whereas BH identifies fewer than $10\%$ of the members.
These results indicate that a collection can exhibit a detectable distributional departure even though its document-level evidence identifies only a small fraction of the training members.

\subsection{Detecting Synthetic Text in Fine-Tuning Data}
\label{sec:exp_cosmopedia}
\label{sec:exp_ft}

We evaluate fine-tuned Pythia-2.8B on Cosmopedia to determine whether synthetic documents were used during fine-tuning.
The target model is trained on $10{,}000$ documents for ten epochs, using the first $512$ tokens of each document.
Each document is scored as a single unit without chunk aggregation.

We compare our method with Ref-Difference, Ref-Ratio, ICP-MIA-SP, and ICP-MIA-Ref.
Ref-Difference measures the reduction in mean negative log-likelihood after fine-tuning, whereas Ref-Ratio uses the negative ratio of the fine-tuned model’s mean negative log-likelihood to that of the initial model.
The ICP-MIA variants measure changes in document likelihood after adding context: ICP-MIA-SP constructs context from masked versions of the document, while ICP-MIA-Ref retrieves semantically similar texts from an external dataset.
As shown in Table~\ref{tab:cosmopedia_auc}, our method outperforms these four baselines across the reported metrics.
Relative to Ref-Ratio, the best-performing baseline, our method improves AUC and TPR@$1\%$ FPR by $4.11$ and $4.38$ percentage points, respectively.
These gains support the usefulness of combining the available membership evidence statistics in document-level discrimination.

\input{web_results}

\begin{figure}[!htbp]
    \setlength{\parskip}{0pt}
    \captionsetup{skip=4pt}
    \captionsetup[subfigure]{skip=2pt}
    \centering
    \begin{subfigure}[t]{0.49\linewidth}
        \centering
        \includegraphics[width=\linewidth]{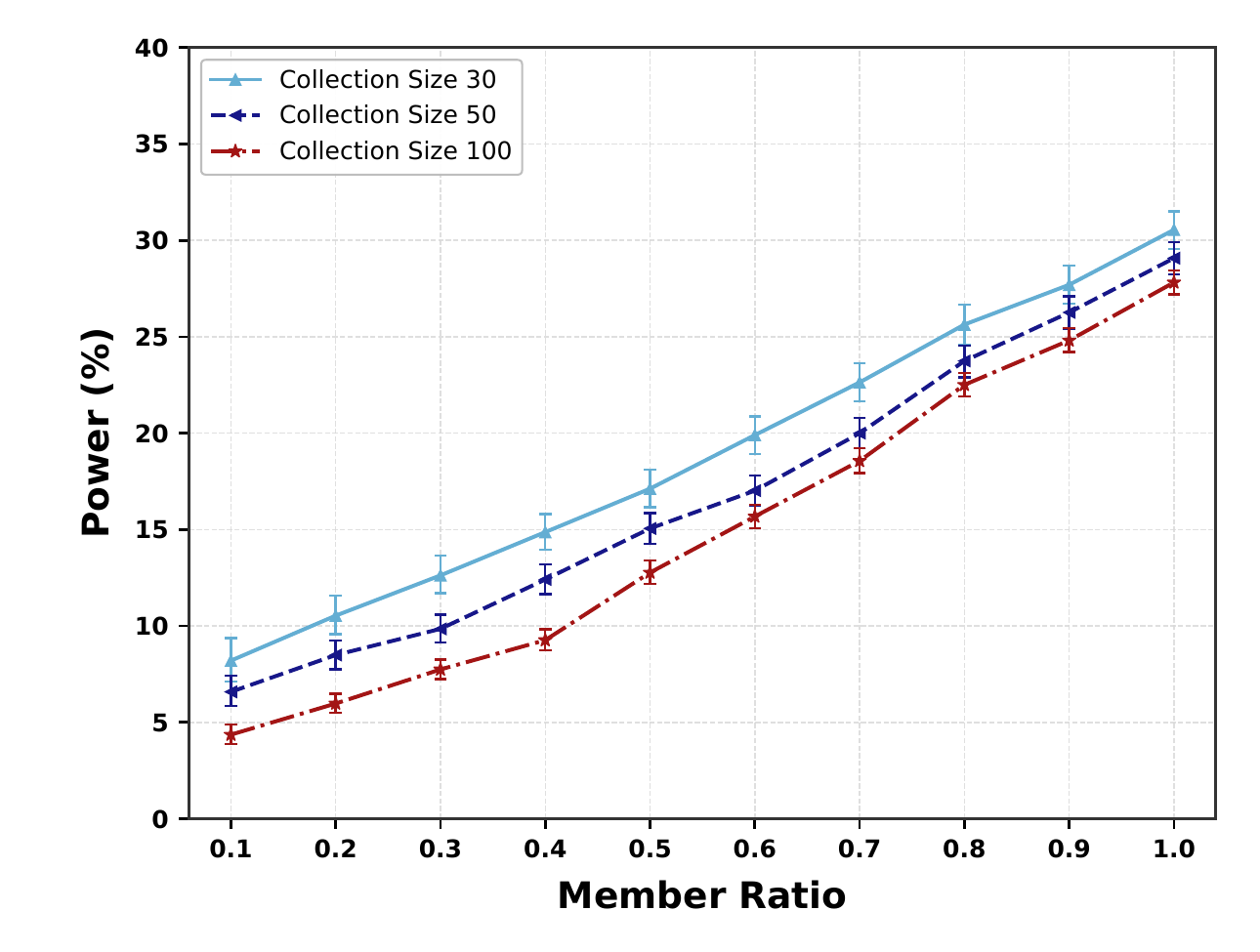}
        \caption{Power}
        \label{fig:cosmopedia_subset_recall}
    \end{subfigure}
    \hfill
    \begin{subfigure}[t]{0.49\linewidth}
        \centering
        \includegraphics[width=\linewidth]{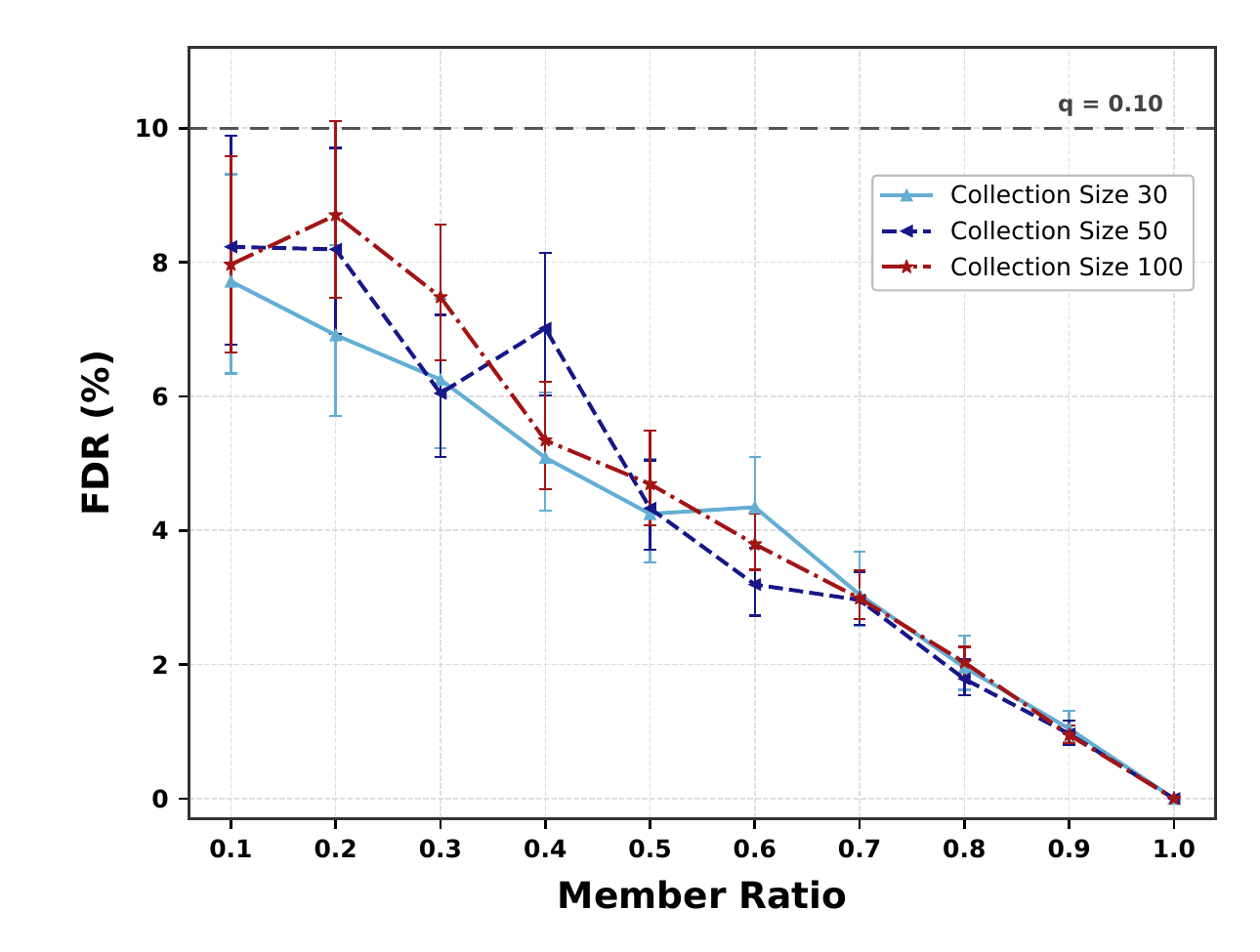}
        \caption{FDR}
        \label{fig:cosmopedia_subset_fdr}
    \end{subfigure}
    \caption{BH subset identification on Cosmopedia. Power and empirical FDR are the respective averages of the within-collection true positive fraction and false discovery proportion over $1{,}000$ collections per configuration. Error bars show pointwise $95\%$ BCa bootstrap confidence intervals. The dashed line marks the target FDR level.}
    \label{fig:cosmopedia_subset}
\end{figure}

At the collection level, Energy rejection rates increase from $9.1$–$26.2\%$ at a $10\%$ member ratio to $96.7$–$100\%$ at a $50\%$ member ratio across the evaluated collection sizes (Figure~\ref{fig:cosmopedia_global}), with larger collections generally achieving higher rejection rates at the same member ratio.
These results indicate that evidence distributed across documents can support reliable detection of a collection-level departure in this fine-tuning setting.
The BH results in Figure~\ref{fig:cosmopedia_subset} reveal a pronounced separation of subset identification.
At $q=0.10$, power increases with the member ratio but stays below one third throughout, while empirical FDR stays below the target $q=0.10$. 
This becomes more pronounced at a $50\%$ member ratio, where the audit can flag nearly all collections while identifying only a limited subset of their members.
This gap makes global detection an incomplete measure of the extent of subset identification.

\subsection{Detecting Medical QA Records in Fine-Tuning Data}
\label{sec:exp_hcm}

We evaluate fine-tuned Llama-3.2-3B-Instruct on HealthcareMagic to audit the use of medical question–answer records during domain-specific adaptation.
We use the checkpoint after the second epoch of fine-tuning on $89{,}732$ records.
Each complete record is an auditing unit, with the instruction and patient question providing context for computing membership evidence from the answer.
We compare with the same four baselines as in \secref{sec:exp_cosmopedia}.
Our method achieves the strongest results across the reported record-level metrics (Table~\ref{tab:hcm_auc}). Its advantage is particularly pronounced under the strict false-positive budget: TPR@$1\%$ FPR exceeds that of ICP-MIA-SP, the strongest baseline here, by $24.11$ percentage points. The strongest baseline changes from Ref-Ratio on Cosmopedia to ICP-MIA-SP on HCM, indicating that reference-model comparisons and context-based evidence have different relative usefulness across these settings.

\input{hcm_results}

\begin{figure}[!htbp]
    \setlength{\parskip}{0pt}
    \captionsetup{skip=4pt}
    \captionsetup[subfigure]{skip=2pt}
    \centering
    \begin{subfigure}[t]{0.49\linewidth}
        \centering
        \includegraphics[width=\linewidth]{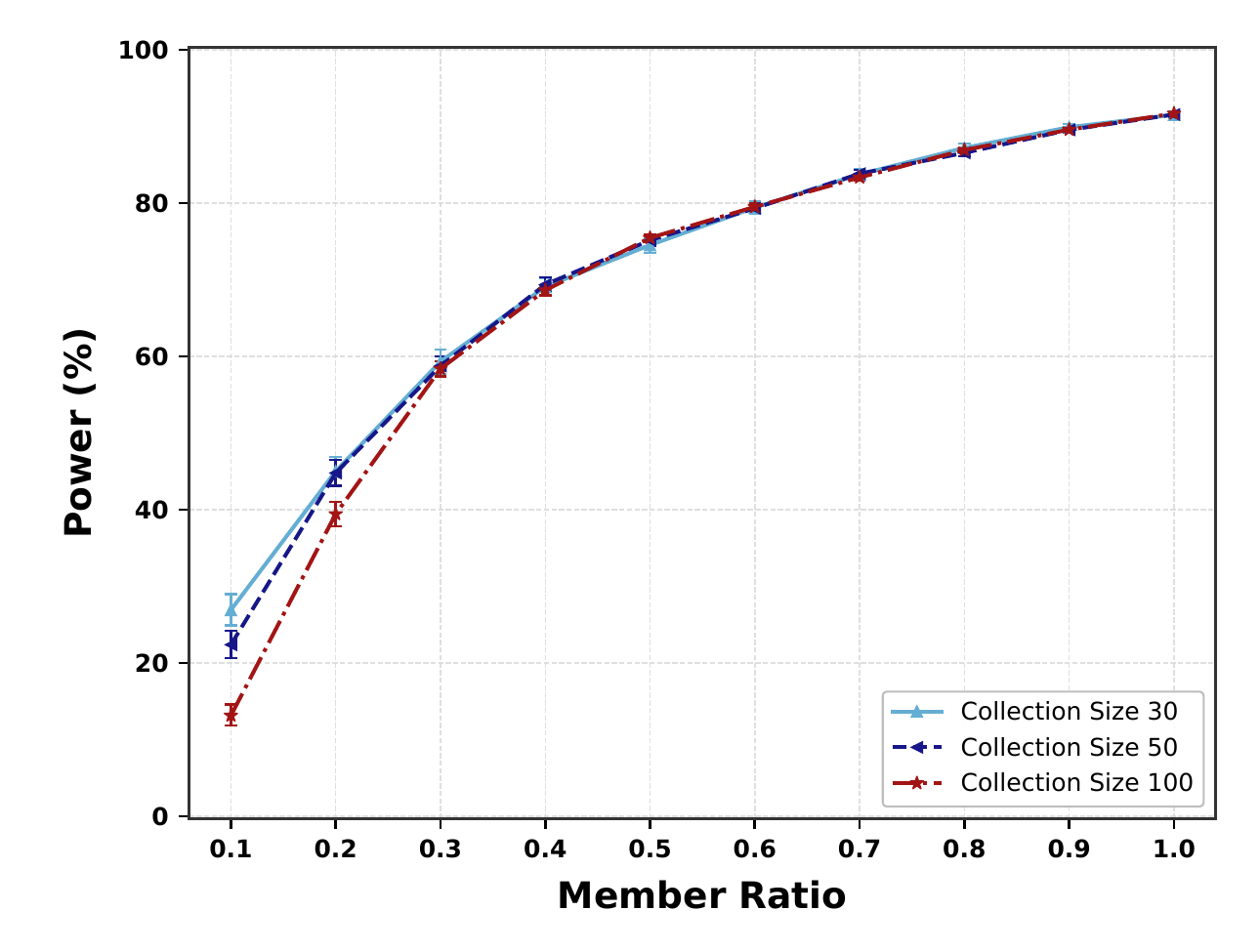}
        \caption{Power}
        \label{fig:hcm_subset_recall}
    \end{subfigure}
    \hfill
    \begin{subfigure}[t]{0.49\linewidth}
        \centering
        \includegraphics[width=\linewidth]{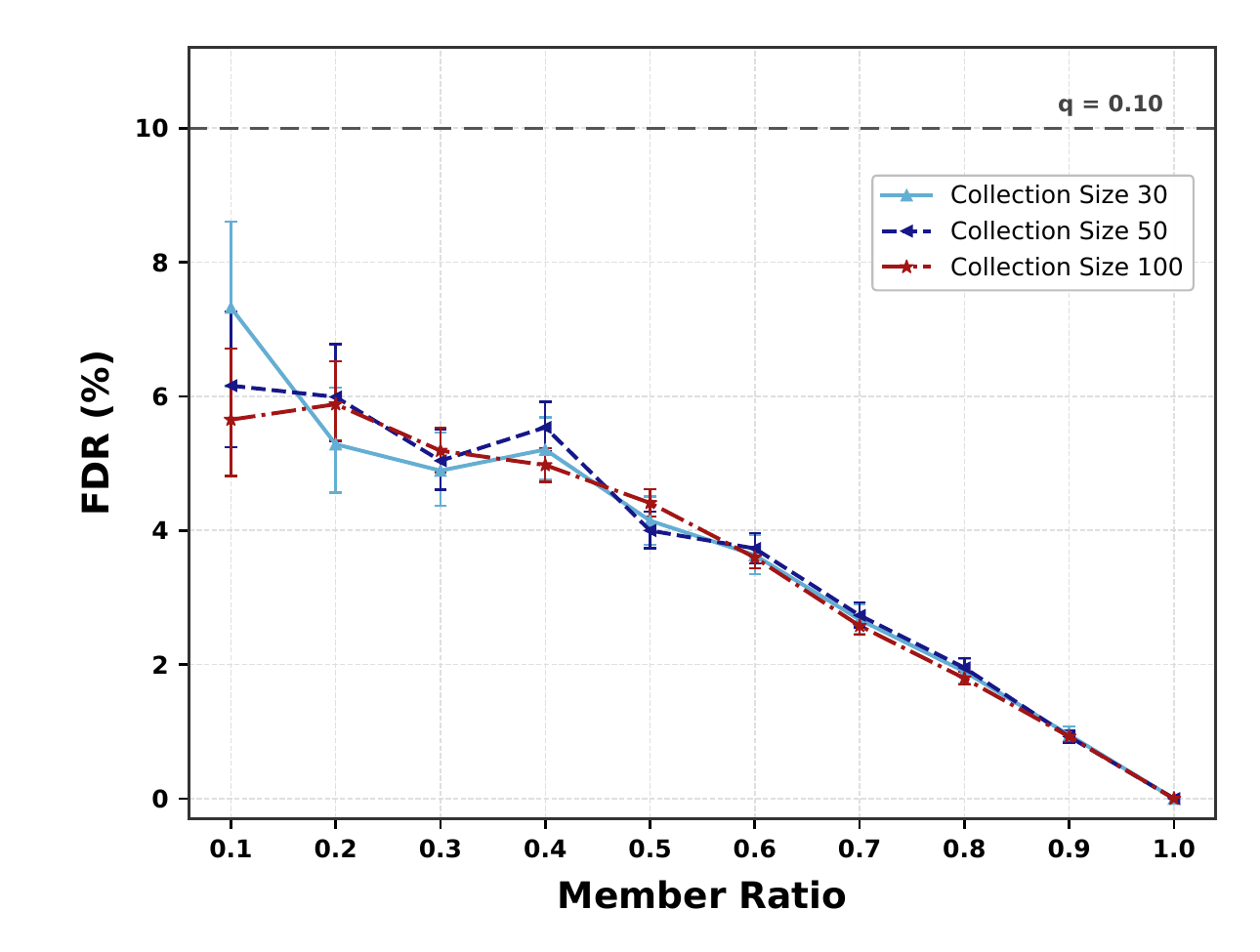}
        \caption{FDR}
        \label{fig:hcm_subset_fdr}
    \end{subfigure}
    \caption{BH subset identification on HCM. Power and empirical FDR are the respective averages of the within-collection true positive fraction and false discovery proportion over $1{,}000$ collections per configuration. Error bars show pointwise $95\%$ BCa bootstrap confidence intervals. The dashed line marks the target FDR level.}
    \label{fig:hcm_subset}
\end{figure}

The stronger individual discrimination is accompanied by more extensive collection-level auditing performance. At a $50\%$ member ratio, the Energy test achieves a $100\%$ rejection rate across all evaluated collection sizes (Figure~\ref{fig:hcm_global}), while BH identifies approximately $75\%$ of members (Figure~\ref{fig:hcm_subset}).
In particular, power increases further with the member ratio, exceeding $90\%$, while empirical FDR remains below the target $q=0.10$. 
Thus, the stronger record-level separation in this setting is accompanied by more extensive member recovery in collection-level auditing.
Compared with Cosmopedia, HCM therefore supports both strong collection-level detection and substantially more complete individual identification at the same target FDR level.

To examine calibration robustness, we vary the size and answer-length composition of the HCM calibration collection while keeping document scores fixed (Supplementary Section~S7).
At $m=30$, the $1{,}000$-, $5{,}000$-, and $10{,}000$-document conditions give similar BH performance over much of the member-ratio range.
Increasing the proportion of long calibration answers raises both power and empirical FDR, with FDR exceeding $q=0.10$ at a $10\%$ member ratio in the long-answer-heavy condition.

\subsection{Summary and Implications}

We summarize the experimental findings and their implications for practical training-data auditing.

\textbf{Identifying specific training documents with controlled errors requires more refined statistical inference than detecting a collection-level departure.}
The high rejection rates for collections containing members and the member subsets recovered by BH at the prespecified FDR levels illustrate the framework's ability to address both objectives.
Even when recovery is incomplete, identifying a subset of documents with strong membership evidence can provide statistical support and concrete leads for investigating training-data use in copyright disputes.

\textbf{In the cases examined, post-training data exhibit stronger document-level membership signals than pretraining data.}
Document-level discrimination is higher on Cosmopedia and HCM than on Pile--arXiv.
One explanation is the extent of training exposure: documents used in relatively few parameter updates may leave weak membership signals, whereas repeated use during post-training may strengthen the model's fit to those documents and increase their separation from non-members.
The Cosmopedia and HCM experiments use checkpoints after ten and two fine-tuning epochs, respectively, consistent with this interpretation.
These findings suggest that auditing performance should be interpreted in relation to training exposure, since limited identification may reflect weak observable signals left by the target documents.

\textbf{Interpreting a distributional departure as evidence of training use requires a suitable non-member comparison population.}
Calibration documents should represent target non-members, since differences in topic, time, or source can affect scores without training exposure.
The error-control guarantees rely on the respective conditional exchangeability assumptions and a scoring rule fixed before document calibration.
Documenting calibration-data provenance and examining sensitivity to plausible non-member populations help assess whether these conditions support the audit conclusions.

\textbf{The reported performance describes auditing with specified model access and auxiliary data.}
Deployment requires the model outputs and any reference models used by the chosen evidence statistics, auxiliary membership labels for fitting, and held-out non-members for calibration.
Within each experimental configuration, we vary target samples while fixing the scoring rule, calibration collection, and evaluation pools.
Higher-quality reference data and more reliable membership labels may further improve auditing performance.

%% file: arxiv_results.tex
\begin{figure}[!htbp]
    \centering
    \renewcommand{\baselinestretch}{1.1}\selectfont
    \captionsetup{font=small,skip=7pt}
    \newsavebox{\arxivresultstablebox}
    \sbox{\arxivresultstablebox}{%
    \footnotesize
    \renewcommand{\arraystretch}{1.15}
    \setlength{\tabcolsep}{4pt}
    \begin{tabular}{@{}lccc@{}}
        \toprule
        \multirow{2}{*}{\textbf{Method}} & \multicolumn{2}{c}{\textbf{TPR@FPR}}
        & \multirow{2}{*}{\textbf{AUC}} \\
        \cmidrule(lr){2-3}
        & $1\%$ & $5\%$ & \\
        \midrule
        PPL & 4.20 & 8.20 & 53.42 \\
        Min-$K$ $20\%$ & 3.80 & 8.40 & 54.43 \\
        Mean-evidence & \underline{7.40} & \underline{19.80} & \textbf{76.04} \\
        Mann--Whitney U & 6.40 & 17.20 & \underline{75.54} \\
        Cauchy (OURS) & \textbf{15.60} & \textbf{23.00} & \underline{75.54} \\
        \bottomrule
    \end{tabular}%
    }
    \newsavebox{\arxivresultsplotbox}
    \sbox{\arxivresultsplotbox}{\includegraphics[width=1.1\wd\arxivresultstablebox]{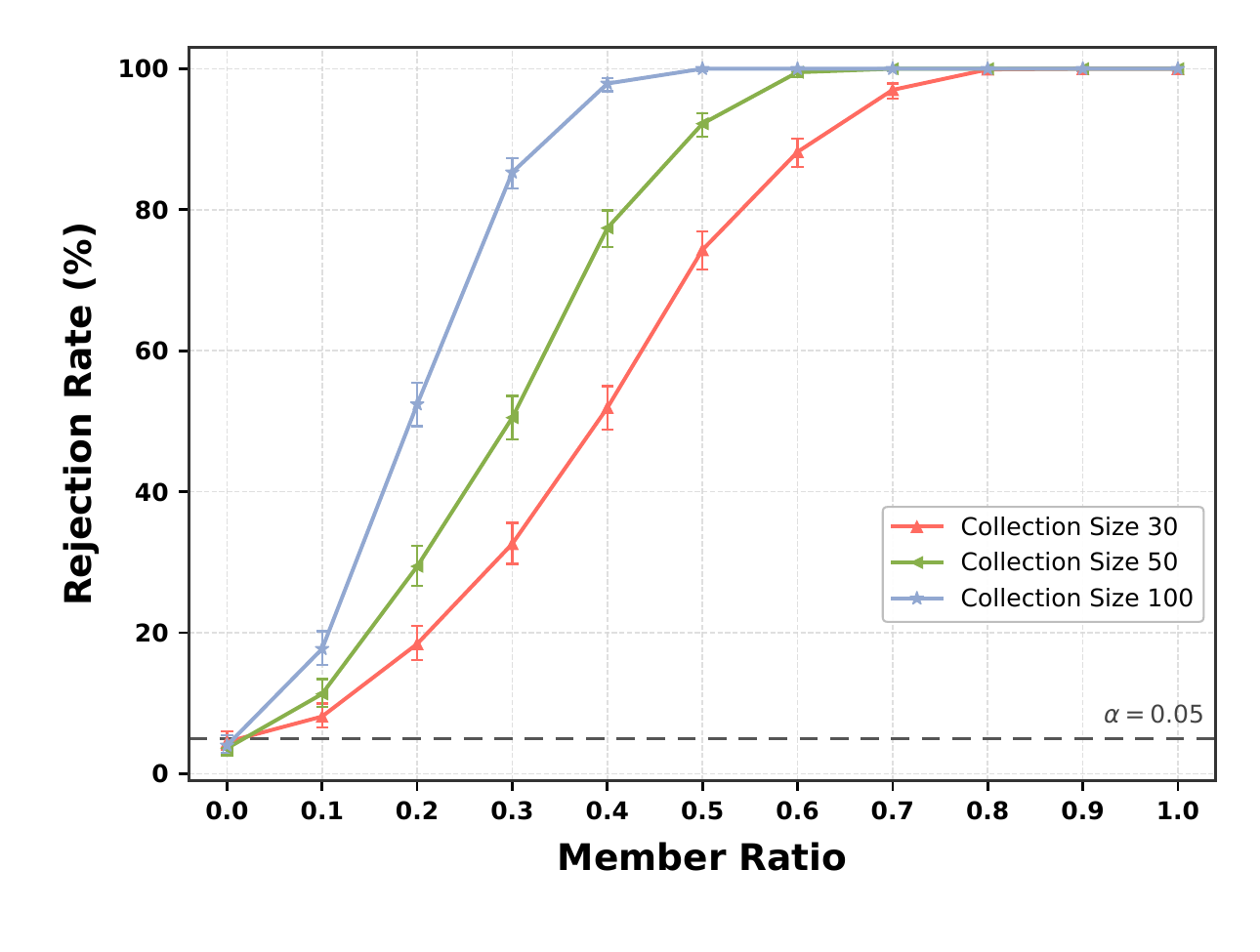}}
    \newlength{\arxivresultsheight}
    \setlength{\arxivresultsheight}{\dimexpr\ht\arxivresultstablebox+\dp\arxivresultstablebox\relax}
    \ifdim\dimexpr\ht\arxivresultsplotbox+\dp\arxivresultsplotbox\relax>\arxivresultsheight
        \setlength{\arxivresultsheight}{\dimexpr\ht\arxivresultsplotbox+\dp\arxivresultsplotbox\relax}
    \fi
    \begin{minipage}[t]{0.49\linewidth}
        \vspace{0pt}
        \begin{minipage}[c][\arxivresultsheight][t]{\linewidth}
            \vspace*{7.4pt}
            \centering
            \usebox{\arxivresultstablebox}
        \end{minipage}
        \captionof{table}{Paper-level discrimination (\%) on Pile–arXiv with Pythia-6.9B. Best in bold; second best underlined.}
        \label{tab:arxiv_auc}
    \end{minipage}
    \hfill
    \begin{minipage}[t]{0.49\linewidth}
        \vspace{0pt}
        \begin{minipage}[c][\arxivresultsheight][t]{\linewidth}
            \vspace{0pt}
            \centering
            \usebox{\arxivresultsplotbox}
        \end{minipage}
        \captionof{figure}{Rejection rates of the energy-based test on Pile–arXiv. Error bars show pointwise $95\%$ Wilson confidence intervals. The dashed line marks the nominal significance level of $5\%$.}
        \label{fig:arxiv_global}
    \end{minipage}
\end{figure}

%% file: web_results.tex
\begin{figure}[!htbp]
    \centering
    \renewcommand{\baselinestretch}{1.1}\selectfont
    \captionsetup{font=small,skip=7pt}
    \newsavebox{\cosmopediaresultstablebox}
    \sbox{\cosmopediaresultstablebox}{%
    \footnotesize
    \renewcommand{\arraystretch}{1.15}
    \setlength{\tabcolsep}{4pt}
    \begin{tabular}{@{}lccc@{}}
        \toprule
        \multirow{2}{*}{\textbf{Method}} & \multicolumn{2}{c}{\textbf{TPR@FPR}}
        & \multirow{2}{*}{\textbf{AUC}} \\
        \cmidrule(lr){2-3}
        & $1\%$ & $5\%$ & \\
        \midrule
        Ref-Difference & 9.89 & 27.02 & 75.65 \\
        Ref-Ratio & \underline{10.82} & \underline{30.04} & \underline{80.69} \\
        ICP-MIA-SP & 2.59 & 9.56 & 60.96 \\
        ICP-MIA-Ref & 2.69 & 10.15 & 62.25 \\
        OURS & \textbf{15.20} & \textbf{39.00} & \textbf{84.80} \\
        \bottomrule
    \end{tabular}%
    }
    \newsavebox{\cosmopediaresultsplotbox}
    \sbox{\cosmopediaresultsplotbox}{\includegraphics[width=1.1\wd\cosmopediaresultstablebox]{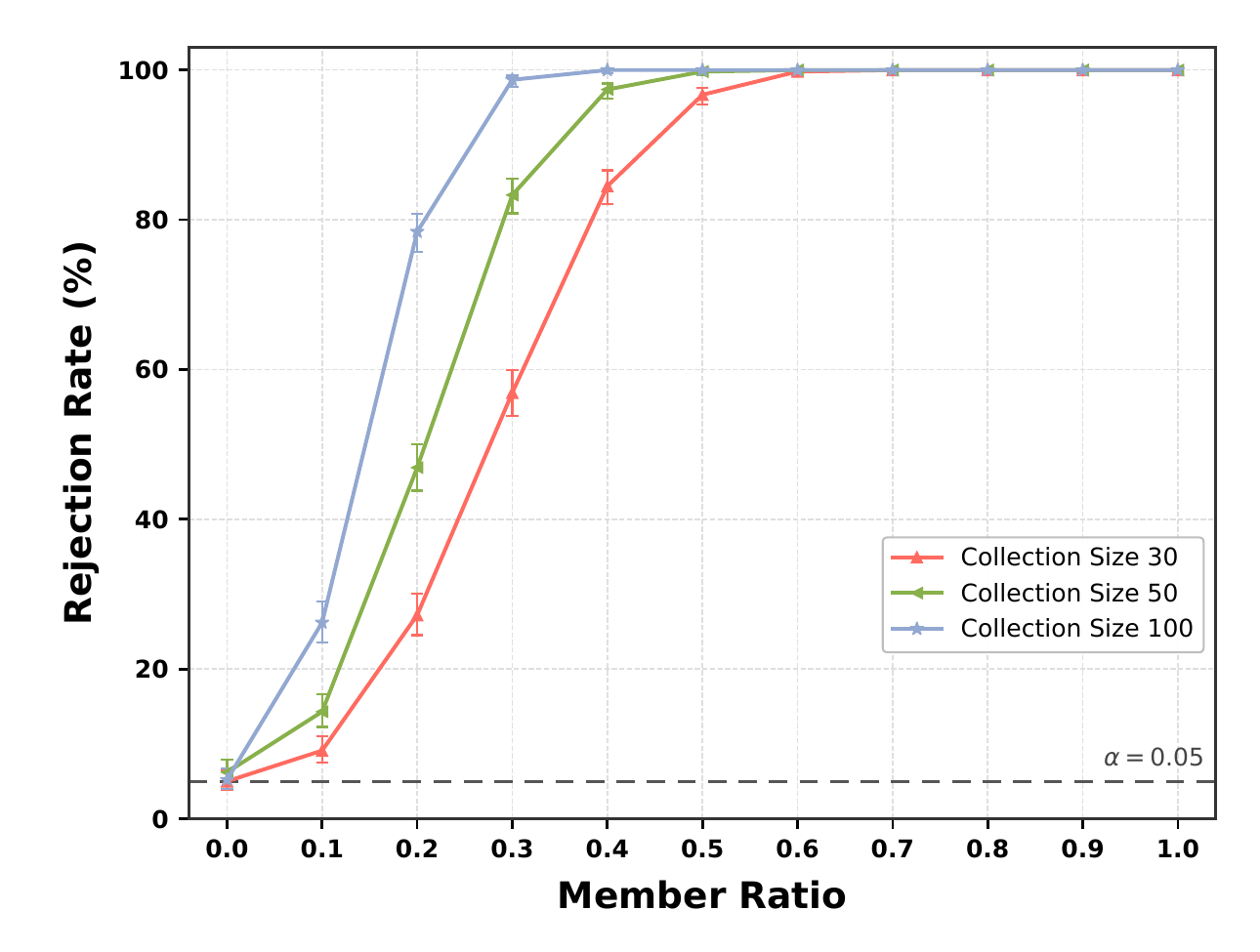}}
    \newlength{\cosmopediaresultsheight}
    \setlength{\cosmopediaresultsheight}{\dimexpr\ht\cosmopediaresultstablebox+\dp\cosmopediaresultstablebox\relax}
    \ifdim\dimexpr\ht\cosmopediaresultsplotbox+\dp\cosmopediaresultsplotbox\relax>\cosmopediaresultsheight
        \setlength{\cosmopediaresultsheight}{\dimexpr\ht\cosmopediaresultsplotbox+\dp\cosmopediaresultsplotbox\relax}
    \fi
    \begin{minipage}[t]{0.49\linewidth}
        \vspace{0pt}
        \begin{minipage}[c][\cosmopediaresultsheight][c]{\linewidth}
            \centering
            \usebox{\cosmopediaresultstablebox}
        \end{minipage}
        \captionof{table}{Document-level discrimination (\%) on Cosmopedia with fine-tuned Pythia-2.8B. Best in bold; second best underlined.}
        \label{tab:cosmopedia_auc}
    \end{minipage}
    \hfill
    \begin{minipage}[t]{0.49\linewidth}
        \vspace{0pt}
        \begin{minipage}[c][\cosmopediaresultsheight][c]{\linewidth}
            \centering
            \usebox{\cosmopediaresultsplotbox}
        \end{minipage}
        \captionof{figure}{Rejection rates of the energy-based test on Cosmopedia. Error bars show pointwise $95\%$ Wilson confidence intervals. The dashed line marks the nominal significance level of $5\%$.}
        \label{fig:cosmopedia_global}
    \end{minipage}
\end{figure}

%% file: hcm_results.tex
\begin{figure}[!htbp]
    \centering
    \renewcommand{\baselinestretch}{1.1}\selectfont
    \captionsetup{font=small,skip=7pt}
    \newsavebox{\hcmresultstablebox}
    \sbox{\hcmresultstablebox}{%
    \footnotesize
    \renewcommand{\arraystretch}{1.15}
    \setlength{\tabcolsep}{4pt}
    \begin{tabular}{@{}lccc@{}}
        \toprule
        \multirow{2}{*}{\textbf{Method}} & \multicolumn{2}{c}{\textbf{TPR@FPR}}
        & \multirow{2}{*}{\textbf{AUC}} \\
        \cmidrule(lr){2-3}
        & $1\%$ & $5\%$ & \\
        \midrule
        Ref-Difference & 1.52 & 9.95 & 76.11 \\
        Ref-Ratio & 1.46 & 10.02 & 78.12 \\
        ICP-MIA-Ref & 10.07 & 33.48 & 83.26 \\
        ICP-MIA-SP & \underline{29.34} & \underline{67.56} & \underline{93.48} \\
        OURS & \textbf{53.45} & \textbf{82.11} & \textbf{97.00} \\
        \bottomrule
    \end{tabular}%
    }
    \newsavebox{\hcmresultsplotbox}
    \sbox{\hcmresultsplotbox}{\includegraphics[width=1.1\wd\hcmresultstablebox]{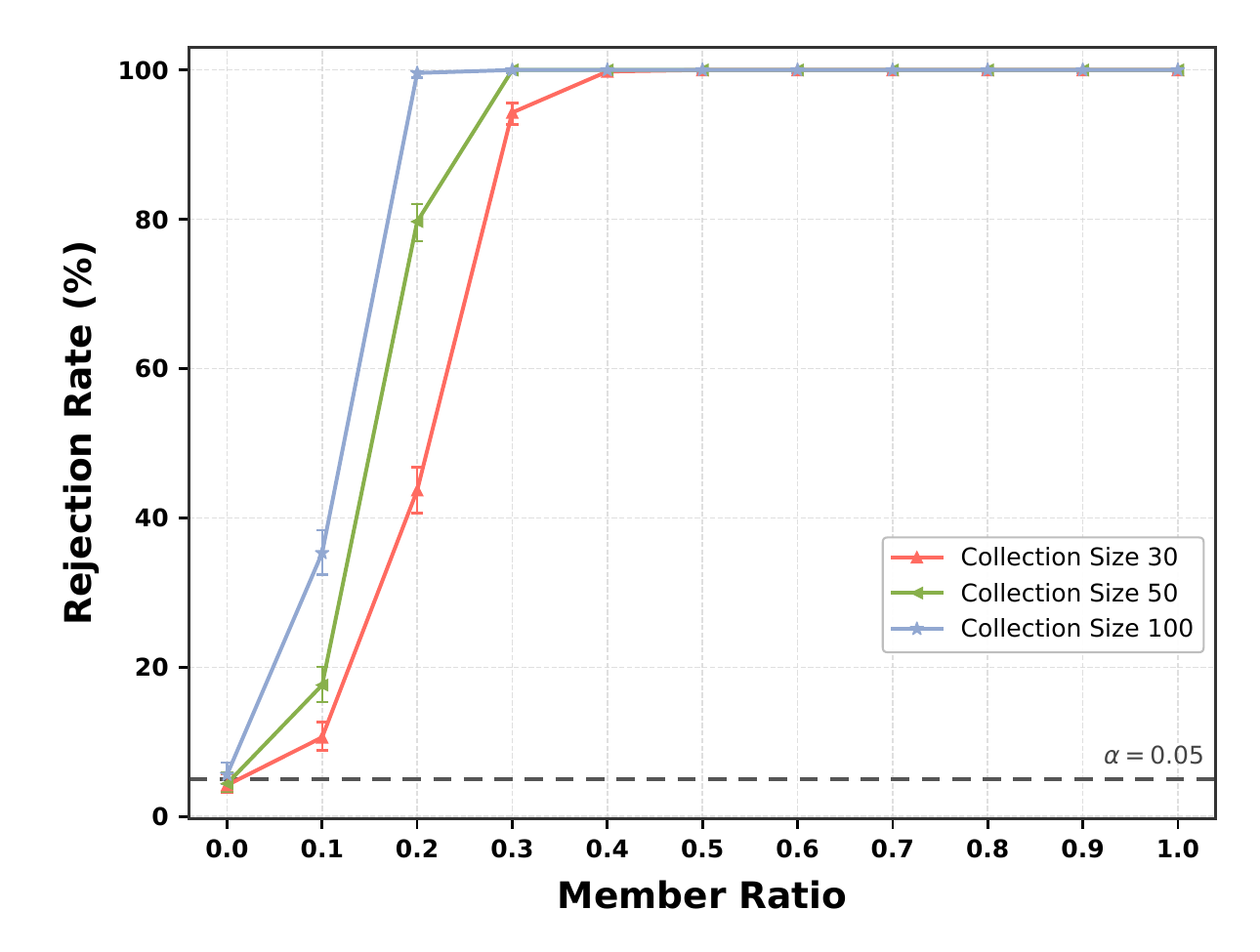}}
    \newlength{\hcmresultsheight}
    \setlength{\hcmresultsheight}{\dimexpr\ht\hcmresultstablebox+\dp\hcmresultstablebox\relax}
    \ifdim\dimexpr\ht\hcmresultsplotbox+\dp\hcmresultsplotbox\relax>\hcmresultsheight
        \setlength{\hcmresultsheight}{\dimexpr\ht\hcmresultsplotbox+\dp\hcmresultsplotbox\relax}
    \fi
    \begin{minipage}[t]{0.49\linewidth}
        \vspace{0pt}
        \begin{minipage}[c][\hcmresultsheight][c]{\linewidth}
            \centering
            \usebox{\hcmresultstablebox}
        \end{minipage}
        \captionof{table}{Record-level discrimination (\%) on HCM with fine-tuned Llama-3.2-3B-Instruct. Best in bold; second best underlined.}
        \label{tab:hcm_auc}
    \end{minipage}
    \hfill
    \begin{minipage}[t]{0.49\linewidth}
        \vspace{0pt}
        \begin{minipage}[c][\hcmresultsheight][c]{\linewidth}
            \centering
            \usebox{\hcmresultsplotbox}
        \end{minipage}
        \captionof{figure}{Rejection rates of the energy-based test on HCM. Error bars show pointwise $95\%$ Wilson confidence intervals. The dashed line marks the nominal significance level of $5\%$.}
        \label{fig:hcm_global}
    \end{minipage}
\end{figure}

%% file: 7_conclusion.tex
\section{Conclusion}\label{sec:conclusion}

Data owners need reliable evidence of whether their documents were used to train LLMs.
A central statistical difficulty is that membership signals are dependent: chunk-level evidence within a document shares context, and document-level scores across a collection share the same fitted evidence map and calibration data, which is a source of dependence that prior auditing methods largely ignore.
The dependence nature of these signals is not well recognized, and existing methods often fail to give reliable guarantees in data-auditing problems.

We develop a framework that remains valid and useful under such dependence: Cauchy combination aggregates dependent chunk-level evidence into document scores, an Energy two-sample test assesses collection-level usage from dependent document scores, and the Benjamini--Hochberg procedure selects suspicious documents with false discovery rate control, all with finite-sample guarantees under conditional exchangeability.
Case studies on books, scientific papers, synthetic text, and medical question--answer records illustrate its use in auditing both pretraining and fine-tuning data.

Our framework uses auxiliary data to construct empirical evidence and held-out non-members to calibrate decisions.
Common MIA approaches use shadow models trained on different datasets~\citep{mia,lira} or held-out data for evidence construction~\citep{recall,maini2024datasetllm}.
For a fixed model, texts newly created and published after release supply non-members; earlier Wikipedia articles and public corpora widely used in LLM training supply candidate members.
Such data are generally easier to obtain than multiple LLM-scale shadow models, whose training is costly and time-consuming.

Several questions merit further study.
How can reference documents be selected or reweighted to account for temporal and domain shifts while retaining error control?
With a limited reference-data budget, how should documents be allocated to improve detection power?
Addressing these questions would help extend reliable auditing to settings with less knowledge of the target model's training data.

%% file: X_appendix.tex
Sections~\ref{app:membership_evidence} and~\ref{app:alt_aggregators} define the membership-evidence statistics and document scoring rules.
Section~\ref{app:theoryproofs} provides the proofs, Section~\ref{app:implementation} records the experimental settings, Section~\ref{sec:exp_booktection} presents the BookTection study, and Section~\ref{app:bonferroni} reports the Bonferroni collection-level results.
Section~\ref{app:hcm_calibration_sensitivity} examines the sensitivity of BH subset identification to the size and distribution of the calibration collection.

\section{Membership Evidence Statistics}
\label{app:membership_evidence}

We define the evidence statistics used in the chunk-level construction of Section~3.1 of the main text.
Let $\mathbf{w}=(w_1,\ldots,w_L)\in\mathcal V^L$ denote a scored token sequence, with vocabulary $\mathcal V$ and prefix $\mathbf{w}_{<t}$ at position $t$.
Each method transforms model outputs, optionally with auxiliary information, into a scalar statistic $e_A(\mathbf{w})$.
Below we list the membership-evidence statistics; Section~\ref{app:evidence_implementation} describes the feature transformations used in the experiments.

\paragraph*{Required model outputs.}
PPL, Zlib, Lowercase, Min-$K\%$, and reference-model comparison use observed-token probabilities or log-probabilities from the relevant model queries.
Min-$K\%++$ additionally requires the full next-token predictive distribution over the vocabulary at each prefix or the corresponding predictive moments.

\paragraph*{Perplexity (PPL).}
Perplexity summarizes how well the target model predicts a token sequence and has been used as a membership-evidence statistic~\citep{reference}.
The evidence statistic is the exponential of the mean negative log-likelihood:
\begin{equation}
e_{\mathrm{PPL}}(\mathbf{w};\theta_{\mathrm{target}})
:=
\mathrm{PPL}(\mathbf{w};\theta_{\mathrm{target}})
=
\exp\left\{-\frac{1}{L}\sum_{t=1}^{L}\log p_{\theta_{\mathrm{target}}}(w_t\mid\mathbf{w}_{<t})\right\}.
\end{equation}
Smaller values indicate stronger membership evidence because the target model assigns higher probabilities to the observed tokens on average.
Text difficulty can confound this comparison: predictable non-member text may have low perplexity, while difficult member text may have high perplexity~\citep{soft}.

\paragraph*{Min-$K\%$.}
Min-$K\%$~\citep{mink} focuses on the tokens that the target model finds least predictable.
For a prescribed percentage $\rho\in(0,100]$, let $\mathcal T_\rho(\mathbf{w};\theta_{\mathrm{target}})$ contain the indices of the bottom $\rho\%$ of the values $\log p_{\theta_{\mathrm{target}}}(w_t\mid\mathbf{w}_{<t})$.
Write $n_\rho:=|\mathcal T_\rho(\mathbf{w};\theta_{\mathrm{target}})|\geq1$.
The resulting membership evidence is
\begin{equation}
e_{\mathrm{Min}\text{-}K,\rho}(\mathbf{w};\theta_{\mathrm{target}})
:=
\frac{1}{n_\rho}
\sum_{t\in\mathcal T_\rho(\mathbf{w};\theta_{\mathrm{target}})}
\log p_{\theta_{\mathrm{target}}}(w_t\mid\mathbf{w}_{<t}).
\end{equation}
Restricting the average to low-probability tokens prevents easy, common tokens from dominating the evidence.
Training exposure may increase the probabilities of even the least-predictable tokens, so larger, less negative values indicate stronger membership evidence.

\paragraph*{Min-$K\%++$.}
Min-$K\%++$~\citep{mink+} extends Min-$K\%$ by standardizing each observed-token log-probability relative to the target model's full predictive distribution at the same prefix.
Define
\begin{equation}
\begin{aligned}
\mu_{\theta_{\mathrm{target}},t}
&=
\sum_{v\in\mathcal V}
p_{\theta_{\mathrm{target}}}(v\mid\mathbf{w}_{<t})
\log p_{\theta_{\mathrm{target}}}(v\mid\mathbf{w}_{<t}),
\\
\sigma_{\theta_{\mathrm{target}},t}^{2}
&=
\sum_{v\in\mathcal V}
p_{\theta_{\mathrm{target}}}(v\mid\mathbf{w}_{<t})
\left(
\log p_{\theta_{\mathrm{target}}}(v\mid\mathbf{w}_{<t})-\mu_{\theta_{\mathrm{target}},t}
\right)^2,
\\
z_{\theta_{\mathrm{target}},t}
&=
\frac{
\log p_{\theta_{\mathrm{target}}}(w_t\mid\mathbf{w}_{<t})-\mu_{\theta_{\mathrm{target}},t}
}{\sigma_{\theta_{\mathrm{target}},t}},
\end{aligned}
\end{equation}
where $\mu_{\theta_{\mathrm{target}},t}$ and $\sigma_{\theta_{\mathrm{target}},t}^{2}$ are, respectively, the probability-weighted mean and variance of the next-token log-probability under $p_{\theta_{\mathrm{target}}}(\cdot\mid\mathbf{w}_{<t})$, and $\sigma_{\theta_{\mathrm{target}},t}>0$ is assumed.
The standardized value measures the observed token's log-probability relative to the distribution of log-probabilities at the same prefix.
Let $\mathcal T_\rho^{++}(\mathbf{w};\theta_{\mathrm{target}})$ contain the indices of the $n_\rho$ smallest standardized values $z_{\theta_{\mathrm{target}},t}$.
The resulting membership evidence is
\begin{equation}
e_{\mathrm{Min}\text{-}K++,\rho}(\mathbf{w};\theta_{\mathrm{target}})
:=
\frac{1}{n_\rho}
\sum_{t\in\mathcal T_\rho^{++}(\mathbf{w};\theta_{\mathrm{target}})}
z_{\theta_{\mathrm{target}},t}.
\end{equation}
Larger values indicate that even the tokens with the lowest standardized log-probabilities are relatively well predicted, yielding stronger membership evidence.

\paragraph*{Zlib.}
Compression length provides a model-free measure of textual regularity.
Following \citet{reference}, let $\mathrm{zlib}(\mathbf{w})$ denote the compressed length, measured in bits, of the underlying text using zlib compression~\citep{zlib}.
The zlib-normalized membership evidence is
\begin{equation}
e_{\mathrm{zlib}}(\mathbf{w};\theta_{\mathrm{target}})
:=
\frac{\log \mathrm{PPL}(\mathbf{w};\theta_{\mathrm{target}})}{\mathrm{zlib}(\mathbf{w})}.
\end{equation}
Repeated substrings and other regular patterns tend to reduce compression length, providing a baseline for interpreting low log-perplexity.
Smaller values indicate stronger membership evidence because the target model achieves low log-perplexity relative to the compressed length.

\paragraph*{Lowercase.}
The Lowercase statistic compares the target model's perplexity on the original text with that on its lowercased version~\citep{reference}.
Here, $\operatorname{lower}(\mathbf{w})$ denotes the token sequence obtained by lowercasing the underlying text and then retokenizing it.
The resulting membership evidence is
\begin{equation}
e_{\mathrm{lower}}(\mathbf{w};\theta_{\mathrm{target}})
:=
\frac{\mathrm{PPL}(\mathbf{w};\theta_{\mathrm{target}})}
{\mathrm{PPL}(\operatorname{lower}(\mathbf{w});\theta_{\mathrm{target}})}.
\end{equation}
Training exposure may improve the model's fit to the original casing of a sequence, reducing its perplexity relative to that of the lowercased version.
A smaller ratio therefore indicates stronger membership evidence.

\paragraph*{Reference-model comparison.}
When a reference model $\mF_{\theta_{\mathrm{ref}}}$ is available, the target model's log-perplexity can be normalized by that of the reference model~\citep{reference}.
The resulting membership evidence is
\begin{equation}
e_{\mathrm{ref}}(\mathbf{w};\theta_{\mathrm{target}},\theta_{\mathrm{ref}})
:=
\frac{\log \mathrm{PPL}(\mathbf{w};\theta_{\mathrm{target}})}
{\log \mathrm{PPL}(\mathbf{w};\theta_{\mathrm{ref}})}.
\end{equation}
Both perplexities are evaluated on the same underlying text, with each model using its own tokenizer and scored-token count.
A smaller ratio indicates that the target model fits the text better relative to the reference model and therefore provides stronger membership evidence.
For fine-tuning audits, the initial model can serve as the reference, allowing the statistic to compare the sequence's fit before and after fine-tuning.

\section{Document-Level Scoring Rules and Comparisons}
\label{app:alt_aggregators}

We compare document scoring rules using the same fixed evidence map and non-member base pool $\{c_h^{\mathrm{base}}\}_{h=1}^{n_{\mathrm{base}}}$.
Mean-evidence, the Mann--Whitney U test, and Cauchy combination are compared on Pile--arXiv in the main text; additional comparisons, including Fisher combination, are reported for BookTection in Table~\ref{tab:booktection_auc}.

Mean-evidence and Max-evidence aggregate the combined chunk evidence directly within each document:
\begin{equation}
\label{eq:doc_mean}
T_{\mathrm{mean}}(d_i) = \frac{1}{J(d_i)}\sum_{j=1}^{J(d_i)} e_{\mathrm{com}}(c_j(d_i)),
\end{equation}
\begin{equation}
\label{eq:doc_max}
T_{\mathrm{max}}(d_i) = \max_{1\le j\le J(d_i)} e_{\mathrm{com}}(c_j(d_i)).
\end{equation}
The mean score summarizes evidence across all chunks, whereas the maximum retains the largest local contribution.

For the Mann--Whitney U test, we compare the document's chunk evidence with that of the fixed non-member base pool using
\begin{equation}
\label{eq:doc_mwu}
T_{\mathrm{MWU}}(d_i) = \sum_{j=1}^{J(d_i)}\sum_{h=1}^{n_{\mathrm{base}}}
\mathbb I\{e_{\mathrm{com}}(c_j(d_i)) > e_{\mathrm{com}}(c_h^{\mathrm{base}})\}.
\end{equation}
This score counts target--base comparisons in which the target chunk has larger evidence and is invariant to strictly increasing transformations of $e_{\mathrm{com}}(\cdot)$.

Cauchy and Fisher combination aggregate the empirical chunk-level $p$-values from Equation~(7) of the main text.
The Cauchy score is defined in Equation~(8); the Fisher score sums logarithmic contributions from the same $p$-values:
\begin{equation}
\label{eq:app_fisher}
T_{\mathrm{Fisher}}(d_i) = -2\sum_{j=1}^{J(d_i)}\log p_{ij}^{\mathrm{ck}}.
\end{equation}
The rank floor $p_{ij}^{\mathrm{ck}}\geq 1/(n_{\mathrm{base}}+1)$ makes the Fisher score finite without additional clipping.
Larger values indicate stronger aggregated membership evidence.
Comparing Cauchy with Mean-evidence holds the evidence map fixed and examines the joint effect of chunk calibration and aggregation.

For individual testing and subset identification, a fixed scoring rule is applied identically to target and calibration documents before using Equation~(9).
Experiment-specific scoring and calibration choices are given in Sections~\ref{app:document_scoring} and~\ref{app:baseline_implementation}.

\section{Proofs of the Theoretical Guarantees}\label{app:theoryproofs}

We first establish an empirical-rank lemma, then prove the results for localized chunk-level evidence, document-level calibration, Energy-based global auditing, BH subset identification, and robustness to approximate exchangeability.

\subsection*{Proof of Lemma~\ref{lem:rank}}

\begin{lemma}[Empirical $p$-values under exchangeability]\label{lem:rank}
Let $\mathcal H$ denote the conditioning sigma-field and let $Z=(Z_0,\ldots,Z_n)$, with $n\geq 1$, be a real-valued vector. Define
\begin{equation}
 p(Z)=\frac{1+\sum_{h=1}^n\mathbb I\{Z_h\geq Z_0\}}{n+1}.
\end{equation}
If $Z$ is exchangeable conditional on $\mathcal H$, then
\begin{equation}
 \Pr\{p(Z)\leq\xi\mid\mathcal H\}
 \leq\frac{\lfloor(n+1)\xi\rfloor}{n+1},
 \qquad\xi\in[0,1].
\end{equation}
Equality holds when all coordinates are distinct almost surely. If the conditional distribution is instead within total variation distance $\delta$ of an exchangeable distribution, the bound becomes
\begin{equation}
 \Pr\{p(Z)\leq\xi\mid\mathcal H\}
 \leq\min\left\{1,\frac{\lfloor(n+1)\xi\rfloor}{n+1}+\delta\right\}.
\end{equation}
\end{lemma}

\begin{proof}
Fix a realization of $\mathcal H$ for which the conditional assumptions hold. For a deterministic vector $z=(z_0,\ldots,z_n)$, define the upper rank
\begin{equation}
 r_h(z)=\sum_{a=0}^n\mathbb I\{z_a\geq z_h\}.
\end{equation}
For every $k\in\{0,\ldots,n+1\}$, at most $k$ coordinates have upper rank at most $k$. Hence
\begin{equation}
 \sum_{h=0}^n\mathbb I\{r_h(z)\leq k\}\leq k.
\end{equation}
Under exchangeability, each indicator has the same conditional expectation, so
\begin{equation}
 \Pr\{r_0(Z)\leq k\mid\mathcal H\}
 =\frac1{n+1}\mathbb E\!\left[
   \sum_{h=0}^n\mathbb I\{r_h(Z)\leq k\}
   \,\middle|\,\mathcal H\right]
 \leq\frac{k}{n+1}.
\end{equation}
Taking $k=\lfloor(n+1)\xi\rfloor$ proves the first result. If all coordinates are distinct, their ranks are exactly $1,\ldots,n+1$, so the displayed deterministic count equals $k$.

For the approximate case, apply the exact result under the exchangeable distribution $Q_{\mathcal H}$. The probability of the event $\{p(Z)\leq\xi\}$ changes by at most $\delta$ between the two distributions. Adding this difference and capping at one proves the claim.
\end{proof}

\subsection*{Proof of Proposition~4.1}

\begin{proof}
The function $p\mapsto\cot(\pi p)$ is decreasing on $(0,1)$. For each $j\in S_i$, the conditions $p^{\mathrm{ck}}_{ij}\leq u$ and $\varepsilon\leq u<1/2$ imply $[p^{\mathrm{ck}}_{ij}]_\varepsilon\leq u$. Thus,
\begin{equation}
 \psi_\varepsilon(p^{\mathrm{ck}}_{ij})
 =\cot\bigl(\pi[p^{\mathrm{ck}}_{ij}]_\varepsilon\bigr)
 \geq\cot(\pi u).
\end{equation}
Summing these $|S_i|$ inequalities, adding the assumed bound for $j\notin S_i$, and dividing by $J(d_i)$ proves Equation~(14).

For the bounded evidence average, each term $1-p^{\mathrm{ck}}_{ij}$ lies in $[0,1]$. Changing only the terms indexed by $S_i$ changes their sum by at most $|S_i|$, and their average by at most $|S_i|/J(d_i)$.
\end{proof}

\subsection*{Proof of Theorem~4.3}

\begin{proof}
Conditional on $\mathcal G$, the document scoring rule is fixed. For a non-member target $i\in\mathcal I_0$, define the $(n_{\mathrm{cal}}+1)$-dimensional vector $T_{\mathrm{joint},i}=(T_{\mathrm{agg}}(d_i),T_{\mathrm{agg}}(d_1^{\mathrm{cal}}),\ldots,T_{\mathrm{agg}}(d_{n_{\mathrm{cal}}}^{\mathrm{cal}}))$. Condition~(i) of Assumption~4.2 makes this vector exchangeable conditional on $\mathcal G$ and the member target scores, and hence also conditional on $\mathcal G$ after averaging over those scores. Lemma~\ref{lem:rank} applies with $Z=T_{\mathrm{joint},i}$, $n=n_{\mathrm{cal}}$, and $\mathcal H=\mathcal G$. Here, $p(Z)$ equals the document-level $p$-value in Equation~(9). The lemma gives Equation~(15), including equality in the first bound for distinct scores. Setting $\xi=\alpha$ proves the type-I error guarantee.
\end{proof}

\subsection*{Proof of Theorem~4.4}

\begin{proof}
Conditional on $\mathcal G$, the scoring rule and the transformation used for Energy testing are fixed. Under condition~(ii) of Assumption~4.2, the resulting pooled input vector $W=(W_1,\ldots,W_{m+n_{\mathrm{cal}}})$ is exchangeable.

Let $\mathcal O(W)$ denote the finite orbit of $W$ under all coordinate permutations. Conditional on $\mathcal G$ and $\mathcal O(W)$, the observed vector is uniformly distributed over this orbit. Moreover, a uniform permutation maps every element of the orbit uniformly onto the same orbit. Since the sampled permutations $\pi_1,\ldots,\pi_B$ are independent of the data and of one another, the vectors $W,\pi_1 W,\ldots,\pi_B W$ are independent and uniformly distributed over $\mathcal O(W)$ under this conditioning.

Applying the same Energy statistic, with fixed group sizes $m$ and $n_{\mathrm{cal}}$, therefore gives an exchangeable vector $(\mathcal E_{\mathrm{obs}},\mathcal E_1,\ldots,\mathcal E_B)$ conditional on $\mathcal G$ and $\mathcal O(W)$. The permutation $p$-value in Equation~(12) is its empirical upper-rank $p$-value. Lemma~\ref{lem:rank}, applied with $n=B$, gives
\begin{equation}
\Pr_{H_0^{\mathrm{glob}}}
\!\left(
p_{\mathrm{global}}\leq\alpha
\mid\mathcal G,\mathcal O(W)
\right)
\leq
\frac{\lfloor(B+1)\alpha\rfloor}{B+1}.
\end{equation}
Taking conditional expectations given $\mathcal G$ proves Equation~(16). The upper-rank argument includes repeated permutations and tied statistics through the weak inequality $\mathcal E_b\geq \mathcal E_{\mathrm{obs}}$.
\end{proof}

\subsection*{Proof of Theorem~4.5}

\begin{proof}
Recall that $m_0=|\mathcal I_0|$ and $m_1=|\mathcal I_1|$ count non-member and member targets, with $m_0+m_1=m$. The result is immediate when $m_0=0$, so suppose that $m_0\geq1$. We first express BH as a stopping rule on the ordered pooled scores.

For the proof, attach independent $\operatorname{Unif}(0,1)$ auxiliary variables to all target and calibration observations, independently of the data. Order the observations by increasing score, using the auxiliary variables only to order equal scores. Let $L=m+n_{\mathrm{cal}}$, and write $z_1\leq\cdots\leq z_L$ for the original scores in this order. After removing the first $t$ observations, let $C_t$, $V_t$, and $A_t$ denote the numbers of remaining calibration documents, non-member targets, and member targets, respectively, and put $D_t=V_t+A_t$.

We consider only the boundaries between distinct score values:
\begin{equation}
\mathcal K
=
\{0\}\cup
\{t\in\{1,\ldots,L-1\}:z_t<z_{t+1}\}.
\end{equation}
At each such boundary, the retained observations contain complete groups of equal scores. Consequently, their counts do not depend on the auxiliary ordering. Define
\begin{equation}
\label{eq:bh_proof_stop}
\tau
=
\min\left\{
t\in\mathcal K:
D_t>0,\quad
\frac{m(1+C_t)}{(n_{\mathrm{cal}}+1)D_t}\leq q
\right\},
\end{equation}
with $\tau=L$ if the set is empty.

We show that the targets retained at $\tau$ are exactly the BH discoveries. At any $t\in\mathcal K$ with $D_t>0$, every retained target satisfies $p_i\leq(1+C_t)/(n_{\mathrm{cal}}+1)$. Thus, if the inequality in \eqnref{eq:bh_proof_stop} holds, then $p_{(D_t)}\leq qD_t/m$, and the BH rejection count is at least $D_t$. Conversely, suppose that BH rejects $r>0$ hypotheses. The largest-rank rule includes every target tied at its selected $p$-value cutoff. Since the empirical $p$-value is nonincreasing in the document score, the BH set consists of all targets at or above its smallest selected score. Removing all pooled observations below this score gives a boundary $t\in\mathcal K$ with $D_t=r$ and
\begin{equation}
\frac{1+C_t}{n_{\mathrm{cal}}+1}
=
p_{(r)}
\leq
\frac{qr}{m}.
\end{equation}
Hence this boundary is feasible. Because $D_t$ is nonincreasing in $t$, the first feasible boundary retains the largest feasible number of targets, which equals the BH rejection count. Its retained target set therefore equals $\widehat{\mS}_{\mathrm{BH}}$. When no boundary is feasible, both procedures return the empty set.

We next use condition~(i) of Assumption~4.2 to control the ratio $V_\tau/(1+C_\tau)$. Let $\mathcal H$ be the sigma-field generated by $\mathcal G$, the member target scores together with their auxiliary variables, and the unordered multiset of calibration and non-member target scores together with their auxiliary variables. Conditional on $\mathcal H$, the ordered scores, the member positions, and the set $\mathcal K$ are fixed. Exchangeability implies that the $n_{\mathrm{cal}}$ calibration labels are uniformly distributed among the remaining $n_{\mathrm{cal}}+m_0$ positions.

Let $\Lambda_a$ denote the label of the observation at position $a$, distinguishing calibration documents, non-member targets, and member targets, and define
\begin{equation}
\mathcal F_t
=
\mathcal H\vee\sigma(\Lambda_1,\ldots,\Lambda_t),
\qquad t=0,\ldots,L.
\end{equation}
The remaining counts $C_t,V_t,A_t$ are $\mathcal F_t$-measurable, since the total counts are fixed. The unrevealed calibration and non-member labels remain uniformly distributed over their unrevealed positions.

Consider the adapted process
\begin{equation}
W_t=\frac{V_t}{1+C_t},
\qquad t=0,\ldots,L.
\end{equation}
If the next position is a member target, then $W_{t+1}=W_t$. Otherwise, conditional on $\mathcal F_t$, write $v=V_t$ and $c=C_t$. When $c>0$, the next label is a non-member target with probability $v/(v+c)$ and a calibration document with probability $c/(v+c)$. Therefore,
\begin{equation}
\begin{aligned}
\mathbb E[W_{t+1}\mid\mathcal F_t]
&=
\frac{v}{v+c}\frac{v-1}{c+1}
+
\frac{c}{v+c}\frac{v}{c}\\
&=
\frac{v}{c+1}
=
W_t.
\end{aligned}
\end{equation}
When $c=0$ and $v>0$, removing a non-member target decreases the ratio from $v$ to $v-1$. Thus $(W_t,\mathcal F_t)_{t=0}^{L}$ is a nonnegative supermartingale.

The boundary set $\mathcal K$ is $\mathcal F_0$-measurable, and the stopping condition in \eqnref{eq:bh_proof_stop} depends only on the remaining counts. Hence $\tau$ is a bounded stopping time. Since $0\leq W_t\leq m_0$ and $W_0=m_0/(n_{\mathrm{cal}}+1)$, the optional stopping theorem gives
\begin{equation}
\mathbb E[W_\tau\mid\mathcal H]
\leq
\frac{m_0}{n_{\mathrm{cal}}+1}.
\end{equation}
Taking conditional expectations given $\mathcal G$ yields the same bound conditional on $\mathcal G$.

On $\{\tau<L\}$, the defining inequality for $\tau$ implies
\begin{equation}
\frac{V_\tau}{D_\tau}
\leq
q\frac{n_{\mathrm{cal}}+1}{m}\frac{V_\tau}{1+C_\tau}.
\end{equation}
On $\{\tau=L\}$, all remaining counts are zero. Using the equivalence with BH established above, we obtain
\begin{equation}
\begin{aligned}
\operatorname{FDR}
\!\left(\widehat{\mS}_{\mathrm{BH}}\mid\mathcal G\right)
&=
\mathbb E\!\left[
\frac{V_\tau}{D_\tau\vee1}
\;\middle|\;\mathcal G
\right]\\
&\leq
q\frac{n_{\mathrm{cal}}+1}{m}
\mathbb E[W_\tau\mid\mathcal G]\\
&\leq
q\frac{m_0}{m}.
\end{aligned}
\end{equation}
This proves Equation~(17).
\end{proof}

\subsection*{Proof of Theorem~4.6}

\begin{proof}
Fix a realization of $\mathcal G$ for which the assumed total variation bound holds, and let $\mathbf Z$ collect all target and calibration document scores. Throughout this proof, $\mathsf J$ and $\mathsf J_\star$ denote the conditional laws of $\mathbf Z$ at this realization.

For any measurable function $f$ taking values in $[0,1]$, the identity $f(\mathbf z)=\int_0^1\mathbb I\{f(\mathbf z)>t\}\,dt$ gives
\begin{equation}\label{eq:tv_bounded_function}
\begin{aligned}
\left|\mathbb E_{\mathsf J}f(\mathbf Z)-\mathbb E_{\mathsf J_\star}f(\mathbf Z)\right|
&=
\left|\int_0^1
\left[\mathsf J\{f(\mathbf Z)>t\}-\mathsf J_\star\{f(\mathbf Z)>t\}\right]\,dt\right|\\
&\leq d_{\mathrm{TV}}(\mathsf J,\mathsf J_\star)
\leq\delta.
\end{aligned}
\end{equation}
We apply this inequality to the error criterion of each procedure.

For document-level testing, fix $i\in\mathcal I_0$ and $\xi\in[0,1]$, and take $f(\mathbf z)=\mathbb I\{p_i(\mathbf z)\leq\xi\}$, where $p_i(\mathbf z)$ is computed by Equation~(9). Under condition~(i) of Assumption~4.2 for $\mathsf J_\star$, Theorem~4.3 yields
\begin{equation}
\begin{aligned}
\Pr(p_i\leq\xi\mid\mathcal G)
&\leq
\mathbb E_{\mathsf J_\star}\mathbb I\{p_i(\mathbf Z)\leq\xi\}+\delta\\
&\leq
\frac{\lfloor(n_{\mathrm{cal}}+1)\xi\rfloor}{n_{\mathrm{cal}}+1}+\delta.
\end{aligned}
\notag
\end{equation}

For BH, let $\widehat{\mS}_{\mathrm{BH}}(\mathbf z)$ denote the discovery set obtained from the score vector $\mathbf z$ at level $q$. The same target documents are non-members under $\mathsf J$ and $\mathsf J_\star$, so the false-discovery proportion is the same function under both laws:
\begin{equation}
f_{\mathrm{BH}}(\mathbf z)
=
\frac{|\widehat{\mS}_{\mathrm{BH}}(\mathbf z)\cap\mathcal I_0|}
{|\widehat{\mS}_{\mathrm{BH}}(\mathbf z)|\vee1}
\in[0,1].
\notag
\end{equation}
Applying \eqnref{eq:tv_bounded_function} and Theorem~4.5 under condition~(i) for $\mathsf J_\star$ gives
\begin{equation}
\begin{aligned}
\operatorname{FDR}(\widehat{\mS}_{\mathrm{BH}}\mid\mathcal G)
&=
\mathbb E_{\mathsf J}f_{\mathrm{BH}}(\mathbf Z)\\
&\leq
\mathbb E_{\mathsf J_\star}f_{\mathrm{BH}}(\mathbf Z)+\delta\\
&\leq
\frac{|\mathcal I_0|}{m}q+\delta.
\end{aligned}
\notag
\end{equation}

For Energy testing under $H_0^{\mathrm{glob}}$, let $\Pi=(\pi_1,\ldots,\pi_B)$ contain the sampled permutations, with the common law $\nu$ specified in Theorem~4.4, independently of the scores under both $\mathsf J$ and $\mathsf J_\star$. Using the same fixed score transformation and permutation test, define
\begin{equation}
h_\alpha(\mathbf z)
=
\int
\mathbb I\{p_{\mathrm{global}}(\mathbf z,\boldsymbol\pi)\leq\alpha\}
\,\nu(d\boldsymbol\pi)
\in[0,1].
\notag
\end{equation}
Thus $h_\alpha(\mathbf z)$ is the rejection probability averaged over the permutation randomness for a fixed score vector. Under condition~(ii) for $\mathsf J_\star$, \eqnref{eq:tv_bounded_function} and Theorem~4.4 imply
\begin{equation}
\begin{aligned}
\Pr_{H_0^{\mathrm{glob}}}(\mathcal E_\alpha\mid\mathcal G)
&=
\mathbb E_{\mathsf J}h_\alpha(\mathbf Z)\\
&\leq
\mathbb E_{\mathsf J_\star}h_\alpha(\mathbf Z)+\delta\\
&\leq
\frac{\lfloor(B+1)\alpha\rfloor}{B+1}+\delta
\leq\alpha+\delta.
\end{aligned}
\notag
\end{equation}

Each error criterion is at most one. Capping the three bounds at one proves Equation~(18), Equation~(19), and Equation~(20). Since the assumed total variation bound holds almost surely, these conclusions hold almost surely conditional on $\mathcal G$.
\end{proof}

\begingroup
\clubpenalty=10000
\widowpenalty=10000
\section{Experimental Settings and Implementation}
\label{app:implementation}
\label{app:finetuning_implementation}

\subsection{Data Allocation and Auditing Splits}
\label{app:data_splits}
\label{app:ft_data}
\label{app:score_calibration}

For BookTection, Cosmopedia, and HCM, we follow the preprocessing protocols used in the corresponding benchmarks.
For Pile--arXiv, we construct the member and non-member pools from the first training shard and the official held-out validation and test splits, respectively.
We first restrict both pools to papers containing at least $8{,}192$ tokens under the Pythia tokenizer.
We retain all eligible non-member papers and randomly sample eligible member papers from the first training shard without replacement.
The number of member papers sampled is $500$ fewer than the number of eligible non-member papers, allowing for additional non-member data needed for calibration.
We then exclude papers whose first $12{,}000$ characters contain a \texttt{date:} field followed on the same line by a four-digit number, applying the same rule to both groups.
The final dataset used in our analyses contains $2{,}275$ member papers and $2{,}850$ non-member papers.

Table~\ref{tab:audit_splits} summarizes the auditing splits.
For BookTection and Pile--arXiv, fitting non-members also serve as the base collection.
Fitting and selection of the scoring rule are completed before any evaluation or calibration documents are used.
The calibration and evaluation pools are mutually disjoint and exclude all fitting and base documents.
The calibration collection is shared by empirical document calibration and Energy testing. 

\begin{table}[htbp]
    \centering
    \renewcommand{\baselinestretch}{1.1}\small
    \caption{Document counts for the auditing splits. All data splits use a random seed of $42$. Paired entries report members/non-members; single entries report non-members. NA indicates an unused stage.}
    \label{tab:audit_splits}
    \label{tab:ft_audit_splits}
    \setlength{\tabcolsep}{5pt}
    \renewcommand{\arraystretch}{1.25}
    \begin{tabular}{@{}lcccc@{}}
        \toprule
        \textbf{Dataset} & \textbf{Fitting} & \textbf{Base} & \textbf{Calibration} & \textbf{Evaluation} \\
        \midrule
        Pile--arXiv & $500/500$ & $500$ & $1{,}850$ & $500/500$ \\
        Cosmopedia & $2{,}000/2{,}000$ & NA & $50{,}000$ & $8{,}000/8{,}000$ \\
        HCM & $2{,}000/2{,}000$ & NA & $10{,}000$ & $10{,}000/10{,}000$ \\
        BookTection & $20/20$ & $20$ & NA & $85/40$ \\
        \bottomrule
    \end{tabular}
    \par\smallskip
    \begin{minipage}{\linewidth}
        \footnotesize
        \textit{Notes.} Base documents reuse the fitting non-members.
    \end{minipage}
\end{table}

\subsection{Target Models and Input Processing}
\label{app:ft_training}

Both pretraining studies use Pythia-6.9B as the target model and OpenLLaMA-7B as the reference model.
Each chunk is scored separately, with context restricted to preceding tokens within that chunk.
Pile--arXiv uses eight non-overlapping $1{,}024$-token chunks from the first $8{,}192$ tokens of each paper.

For BookTection, no BOS/EOS tokens are added and scoring starts at the second token.
Reference-model scoring uses its own tokenizer and predicted-token count, with a window of $2{,}048$ tokens and stride $1{,}024$.
The target and reference models are loaded in BF16 and FP16, respectively; token log-probabilities and Min-$K\%++$ moments are computed in FP32.

For the fine-tuning experiments, we update all parameters of \texttt{pythia-2.8b} on Cosmopedia and \texttt{Llama-3.2-3B-Instruct} on HCM, following prior work~\citep{wbc,icpmia}.
Each initial model serves as the reference model; the HCM reference has already undergone instruction tuning.
On Cosmopedia, training uses the first $512$ tokens of each document with the next-token prediction objective, and auditing likelihoods use the same prediction positions.
On HCM, we use the Llama~3 single-turn chat format, with the instruction and question as context; only the response and its end-of-turn token contribute to the training loss and auditing likelihood.
Table~\ref{tab:ft_training_settings} gives the remaining training settings and the audited checkpoints.

\begin{table}[!ht]
    \centering
    \renewcommand{\baselinestretch}{1.1}\small
    \caption{Fine-tuning configurations for Cosmopedia and HCM. Both runs use BF16 precision and random seed $42$.}
    \label{tab:ft_training_settings}
    \setlength{\tabcolsep}{5pt}
    \renewcommand{\arraystretch}{1.15}
    \begin{tabular}{@{}>{\raggedright\arraybackslash}p{0.32\linewidth}>{\raggedright\arraybackslash}p{0.30\linewidth}>{\raggedright\arraybackslash}p{0.30\linewidth}@{}}
        \toprule
        \textbf{Setting} & \textbf{Cosmopedia} & \textbf{HCM} \\
        \midrule
        Training records & $10{,}000$ & $89{,}732$ \\
        Maximum input length & $512$ tokens & $2{,}048$ tokens \\
        Optimizer & AdamW & AdamW \\
        Learning rate & $10^{-5}$ & $2\times10^{-4}$ \\
        Weight decay & $0.1$ & $0$ \\
        Learning-rate schedule & Linear & Cosine \\
        Warmup steps & $500$ & $1{,}683$ \\
        Effective batch size & $48$ & $16$ \\
        Training epochs & $10$ & $3$ \\
        Audited epoch & $10$ & $2$ \\
        Gradient checkpointing & Disabled & Enabled \\
        \bottomrule
    \end{tabular}
\end{table}

\subsection{Evidence Construction and Fitting}
\label{app:evidence_implementation}
\label{app:ft_evidence}

\paragraph*{Implemented feature inputs.}
All four studies use the $18$ feature inputs listed in Table~\ref{tab:ensemble_features}.
Here, $x$ denotes the text input, and $\mathbf w$ denotes its token sequence under the relevant model's tokenizer, as in Section~\ref{app:membership_evidence}.
We write $L_T(x)$ and $L_R(x)$ for the target and reference models' mean NLLs (log-perplexities) in natural-log units, using the scored positions specified in Section~\ref{app:ft_training} and each model's own scored-token count.
Relative to the definitions in Section~\ref{app:membership_evidence}, the implemented inputs use a negative NLL ratio for lowercase comparison and reverse the signs of the ordinary Min-$K\%$ and reference-model ratio scores.

Write $x^{\mathrm{lower}}$ for the lowercased scoring text and $b_{\mathrm{zlib}}(x)$ for the byte length after UTF-8 encoding and zlib compression of $x$.
Using bytes instead of bits makes the implemented compression-normalized input eight times the native statistic for the same scored text.
On Cosmopedia, both features start from the decoded prefix; the lowercase text is retokenized and capped at $512$ tokens, yielding $427$--$512$ tokens.

For the two Min-$K\%$ families, write $\ell_t=\log p_{\theta_{\mathrm{target}}}(w_t\mid\mathbf w_{<t})$ for the target model's log-probability at a scored position.
Let $\widetilde z_t$ be its standardized value from Section~\ref{app:membership_evidence}, with the predictive variance floored at $10^{-8}$.
The sets $\mathcal T_\rho(x)$ and $\mathcal T^{++}_\rho(x)$ select the bottom $\rho\%$ of $\ell_t$ and $\widetilde z_t$, respectively, among the scored positions; $n_\rho(x)$ is the number selected by either rule.
Both families use $\rho\in\{5,10,20,30,40,50,60\}$ and retain at least one token.

\begin{table}[htbp]
    \centering
    \renewcommand{\baselinestretch}{1.1}\small
    \caption{Feature inputs used to fit the combined membership-evidence score in all four studies.}
    \label{tab:ensemble_features}
    \label{tab:ft_cosmopedia_features}
    \setlength{\tabcolsep}{5pt}
    \renewcommand{\arraystretch}{1.6}
    \begin{tabular}{@{}lcl@{}}
        \toprule
        \textbf{Feature group} & \textbf{Count} & \textbf{Implemented input} \\
        \midrule
        Perplexity & $1$ & $\exp\{L_T(x)\}$ \\
        Lowercase comparison & $1$ & $-L_T(x^{\mathrm{lower}})/L_T(x)$ \\
        Compression normalization & $1$ & $L_T(x)/b_{\mathrm{zlib}}(x)$ \\
        Min-$K\%$ & $7$ & $-n_\rho(x)^{-1}\sum_{t\in\mathcal T_\rho(x)}\ell_t$ \\
        Min-$K\%++$ & $7$ & $n_\rho(x)^{-1}\sum_{t\in\mathcal T^{++}_\rho(x)}\widetilde z_t$ \\
        Reference-model ratio & $1$ & $-L_T(x)/L_R(x)$ \\
        \bottomrule
    \end{tabular}
\end{table}

\paragraph*{Fitting configuration.}
We fit a separate evidence map for each dataset, using chunks as fitting observations in the pretraining studies and documents in the fine-tuning studies.
All four pipelines use \texttt{StandardScaler} followed by logistic regression with L2 regularization, $C=1$, \texttt{lbfgs}, at most $1{,}000$ iterations, and \texttt{random\_state=42}.
Only fitting observations enter standardization and classifier fitting.
The scorer returns the raw \texttt{decision\_function} logit as the combined evidence defined in Equation~(6) of the main text.

\subsection{Document Scoring and Calibration}
\label{app:document_scoring}

BookTection and Pile--arXiv apply the chunk calibration and Cauchy aggregation in Equations~(7) and~(8) of the main text, with $\varepsilon=1/[2(n_{\mathrm{base}}+1)]$; clipping affects only the Cauchy transformation input.
BookTection uses the Cauchy score for book ranking without final empirical document calibration.
For the three main-text studies, Table~\ref{tab:scoring_pipeline} identifies the document score and the Energy input.
Each study uses Equation~(9) for empirical document calibration with the calibration collection in Table~\ref{tab:audit_splits}.

\begin{table}[htbp]
    \centering
    \renewcommand{\baselinestretch}{1.1}\small
    \caption{Document scores and Energy inputs in the main-text experiments. Individual testing and BH use empirical document-level $p$-values obtained from the listed document scores.}
    \label{tab:scoring_pipeline}
    \setlength{\tabcolsep}{5pt}
    \renewcommand{\arraystretch}{1.2}
    \begin{tabular}{@{}>{\raggedright\arraybackslash}p{0.18\linewidth}>{\raggedright\arraybackslash}p{0.35\linewidth}>{\raggedright\arraybackslash}p{0.39\linewidth}@{}}
        \toprule
        Dataset & Document score & Energy input \\
        \midrule
        Pile--arXiv & Cauchy aggregation of chunk $p$-values & Fixed bounded transformation in Equation~(10) \\
        Cosmopedia & Document logit, $T_{\mathrm{agg}}(d)=e_{\mathrm{com}}(d)$ & Untransformed document logit \\
        HCM & Document logit, $T_{\mathrm{agg}}(d)=e_{\mathrm{com}}(d)$ & Untransformed document logit \\
        \bottomrule
    \end{tabular}
\end{table}

\subsection{Baseline Configurations}
\label{app:baseline_implementation}

For the pretraining comparisons, PPL and Min-$K\%$ average their chunk statistics from Section~\ref{app:membership_evidence}, with $\rho=20$ for the Pile--arXiv Min-$K\%$ baseline.
Mean-evidence and the other aggregation baselines follow Section~\ref{app:alt_aggregators}.

For Cosmopedia and HCM, Ref-Difference is $L_R(x)-L_T(x)$, where $L_T$ and $L_R$ are the mean NLLs defined in Section~\ref{app:evidence_implementation}.
Ref-Ratio is $-L_T(x)/(L_R(x)+10^{-8})$ on Cosmopedia and $-L_T(x)/L_R(x)$ on HCM.
These baseline scores are larger for more member-like documents.

ICP-MIA-SP and ICP-MIA-Ref~\citep{icpmia} compare the target model's NLL with and without an added context.
SP constructs $20$ candidate contexts by masking $70\%$ of words containing English letters and replacing them with the literal word \texttt{word}.
Ref retrieves $10$ candidate contexts from Dolly-15k using all-MiniLM-L6-v2.
Each candidate is evaluated separately, and the score is the smallest valid context-conditioned NLL minus the NLL without the added context.

For these ICP-MIA baselines, the maximum total input length, including the added context, is $2{,}048$ tokens on Cosmopedia and $4{,}096$ tokens on HCM.
On Cosmopedia, the context is followed by the original $512$ target token IDs, and both NLLs use the same $511$ prediction positions.
Candidates exceeding this input limit are excluded; a document with no valid Ref candidate receives a score of zero.
On HCM, SP masks the response to form its candidate contexts, and both baselines use the response-only scoring rule in Section~\ref{app:ft_training}.

\subsection{Evaluation Protocols}
\label{app:evaluation_protocols}

\paragraph*{Target-collection sampling.}
For Pile--arXiv, Cosmopedia, and HCM, we sample $R=1{,}000$ target collections for each combination of size $m\in\{30,50,100\}$ and member ratio $\eta\in\{0.1,0.2,\ldots,1.0\}$.
Each collection contains $m\eta$ members and $m(1-\eta)$ non-members, sampled without replacement from the fixed evaluation pools reported in Table~\ref{tab:audit_splits}; documents may recur across collections.
For global auditing, we additionally sample $R$ entirely non-member collections for each $m$; these collections are reused across member-ratio settings.

\paragraph*{Test configuration.}
For global auditing, we evaluate the Energy permutation test described in Section~\ref{app:energy_test} at significance level $\alpha=0.05$.
For subset identification, we evaluate the BH procedure separately at $q=0.10$, applying Equation~(13) of the main text to all $m$ empirically calibrated document $p$-values in each target collection.

\paragraph*{Performance summaries.}
The empirical rejection rate of the Energy permutation test is the fraction of the $R$ collections for which $p_{\mathrm{global}}\leq\alpha$, with pointwise $95\%$ Wilson score intervals~\citep{Wilson1927probable}.
For mixed collections, let $\mathrm{TP}_a$, $\mathrm{FP}_a$, and $m_{1,a}$ denote the true discoveries, false discoveries, and number of members in collection $a$.
Empirical power and FDR are
\begin{equation}
\label{eq:implementation_bh_metrics}
\widehat{\mathrm{Power}}
=\frac{1}{R}\sum_{a=1}^{R}\frac{\mathrm{TP}_a}{m_{1,a}},
\qquad
\widehat{\mathrm{FDR}}
=\frac{1}{R}\sum_{a=1}^{R}
\frac{\mathrm{FP}_a}{\max(\mathrm{TP}_a+\mathrm{FP}_a,1)}.
\end{equation}
The false discovery proportion is set to zero for an empty discovery set.
Pointwise $95\%$ confidence intervals for BH power and FDR are constructed using the bias-corrected and accelerated (BCa) bootstrap method~\citep{Efron1987better}, with $9{,}999$ resamples of the $R$ collection-level true positive fractions and false discovery proportions, respectively.
These intervals quantify target-collection sampling uncertainty conditional on the fixed scoring rule, calibration collection, and evaluation pools.
For the calibration-sensitivity comparisons in Section~\ref{app:hcm_calibration_sensitivity}, a separate fixed random stream supplies bootstrap indices shared across conditions at each member ratio.

\subsection{Energy Permutation Test Implementation}
\label{app:energy_test}

For Pile--arXiv, Cosmopedia, and HCM, the Energy permutation test compares each target collection with all calibration documents in Table~\ref{tab:audit_splits}, applying the dataset-specific input rule in Table~\ref{tab:scoring_pipeline} to both groups.
Let $u_i$ and $v_\ell$ denote the resulting scalar target and calibration inputs, respectively.
Across these datasets, we use the same U-centered statistic, which is a positive multiple of
\begin{equation}
\label{eq:energy_centered_statistic}
\begin{aligned}
\widehat{\mathcal E}
={}&\frac{2}{mn_{\mathrm{cal}}}
\sum_{i=1}^{m}\sum_{\ell=1}^{n_{\mathrm{cal}}}|u_i-v_\ell|\\
&-\frac{1}{m(m-1)}
\sum_{\substack{i,i'=1\\i\neq i'}}^{m}|u_i-u_{i'}|\\
&-\frac{1}{n_{\mathrm{cal}}(n_{\mathrm{cal}}-1)}
\sum_{\substack{\ell,\ell'=1\\\ell\neq\ell'}}^{n_{\mathrm{cal}}}|v_\ell-v_{\ell'}|,
\end{aligned}
\end{equation}
where $m,n_{\mathrm{cal}}\geq2$.
The positive factor depends only on the fixed group sizes and therefore leaves permutation $p$-values unchanged.
We retain negative statistic values and apply the permutation rule in Equation~(12) of the main text with $1{,}999$ permutations.
\endgroup

\section{Book-Level Membership Inference on BookTection}
\label{sec:exp_booktection}
We present a book-level case study of document-level individual auditing (Problem 1) on BookTection using Pythia-6.9B~\citep{biderman2023pythia}.
Each book is treated as a document represented by approximately $35$ complete passages from the benchmark’s Medium setting.
We evaluate book-level discrimination and examine how membership evidence is distributed across passages within each book.

We compare Cauchy aggregation with PPL~\citep{loss}, Min-$K$ $20\%$~\citep{mink}, Mean-evidence, the Mann--Whitney U test~\citep{mia_scaling}, and Fisher combination~\citep{hess2007fisher}, with the aggregation rules defined in Section~\ref{app:alt_aggregators}.
As shown in Table~\ref{tab:booktection_auc}, Mean-evidence, the Mann--Whitney U test, Fisher combination, and Cauchy aggregation outperform the individual-indicator baselines across all three reported metrics.
Relative to Mean-evidence, Cauchy aggregation yields a higher AUC ($96.41\%$ versus $95.56\%$) and higher TPRs at both $1\%$ FPR ($80.00\%$ versus $57.65\%$) and $5\%$ FPR ($89.41\%$ versus $84.71\%$).
Using the same empirical chunk-level $p$-values, Cauchy achieves higher AUC and TPR at $1\%$ FPR than Fisher combination, while Fisher yields a slightly higher TPR at $5\%$ FPR ($90.59\%$ versus $89.41\%$).

To examine the distribution of evidence within books, we use the unclipped chunk quantities in Equation~(7) of the main text to define passage flags and their within-book proportion:
\begin{equation}
\label{eq:book_flagged_fraction}
F_{ij}=\mathbb I\{p_{ij}^{\mathrm{ck}}<0.1\},
\qquad R_i=\frac{1}{J(d_i)}\sum_{j=1}^{J(d_i)}F_{ij}.
\end{equation}
These passage flags receive no multiplicity correction and identify locations for further inspection.
Figure~\ref{fig:booktection_flagged_chunks} summarizes $R_i$ using $10$-percentage-point bins for Medium passages, with $100\%$ included in the final bin.
Candidate-member books generally have higher flagged-passage proportions than non-member books.

Table~\ref{tab:book-examples-large} presents the ten books with the highest exact Medium flagged-passage proportions, with ties ordered by decreasing Cauchy score.
Their flagged proportions range from $81.8\%$ to $100\%$.
For each book, we also report the bounded Cauchy-transformed score $u(d)$ defined in Equation~(10), with smaller values indicating stronger aggregated membership evidence.

\input{booktection_results}

\input{booktection_summary}

\section{Bonferroni Collection-Level Auditing}
\label{app:bonferroni}

For problem 2, we also apply Bonferroni correction to the empirical document-level $p$-values in Equation~(9) of the main text.
For a collection of $m$ documents, the rule rejects the intersection of the document non-member nulls when
\begin{equation}
\label{eq:app_bonferroni}
\min_{1\leq i\leq m}p_i\leq\frac{\alpha}{m}.
\end{equation}
Under the document-level validity in Theorem~4.3, the union bound limits the probability of rejecting an entirely non-member collection to $\alpha$, without requiring independent $p$-values.
\textbf{Bonferroni thus requires sufficiently strong evidence from at least one document; Energy tests equality of score distributions and can accumulate evidence across documents.}

\begin{figure}[!htbp]
    \centering
    \captionsetup{skip=4pt}
    \captionsetup[subfigure]{skip=2pt}
    \begin{subfigure}[t]{0.48\linewidth}
        \centering
        \includegraphics[width=\linewidth]{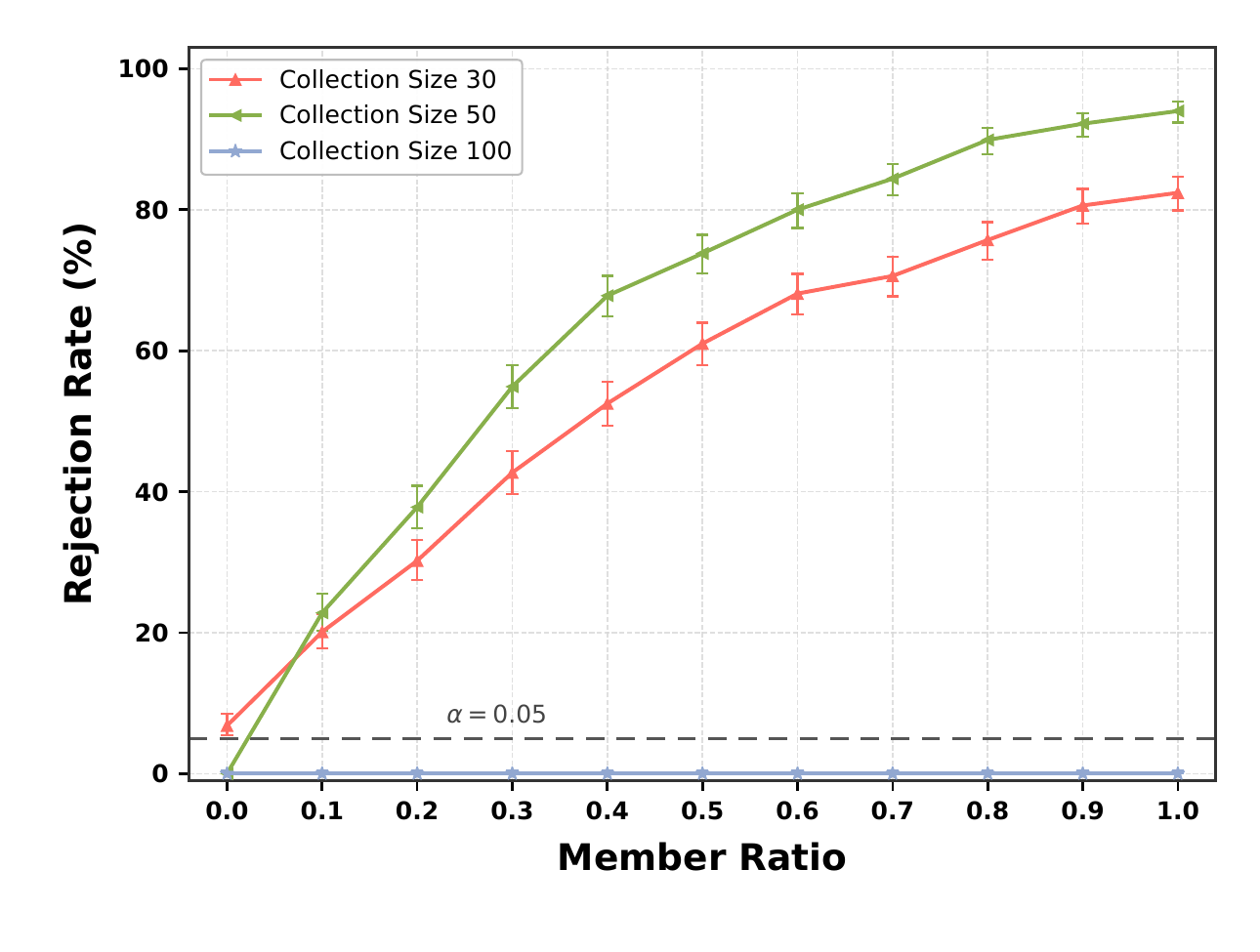}
        \caption{Pile--arXiv}
        \label{fig:arxiv_bonferroni}
    \end{subfigure}
    \hfill
    \begin{subfigure}[t]{0.48\linewidth}
        \centering
        \includegraphics[width=\linewidth]{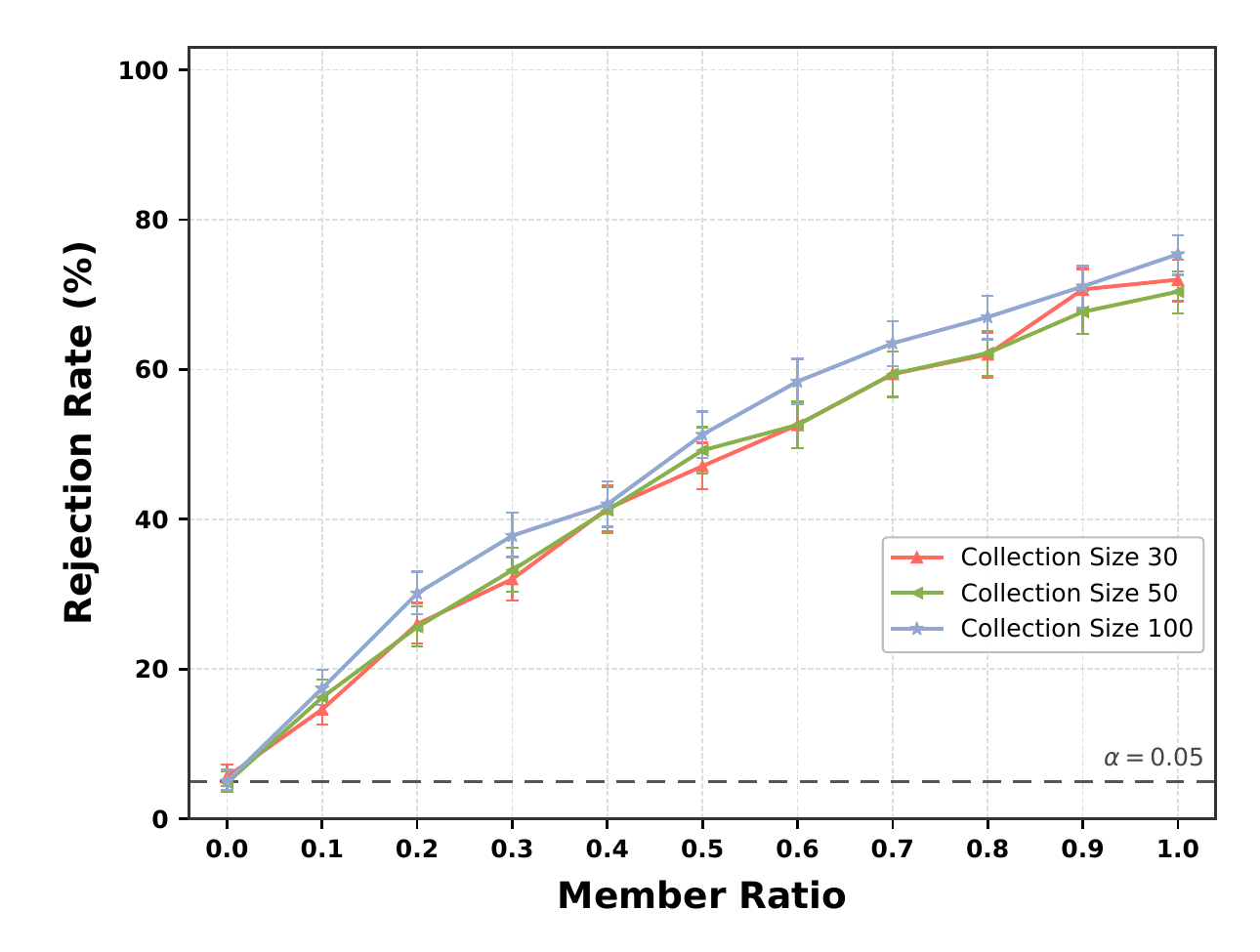}
        \caption{Cosmopedia}
        \label{fig:cosmopedia_bonferroni}
    \end{subfigure}

    \medskip
    \begin{subfigure}[t]{0.48\linewidth}
        \centering
        \includegraphics[width=\linewidth]{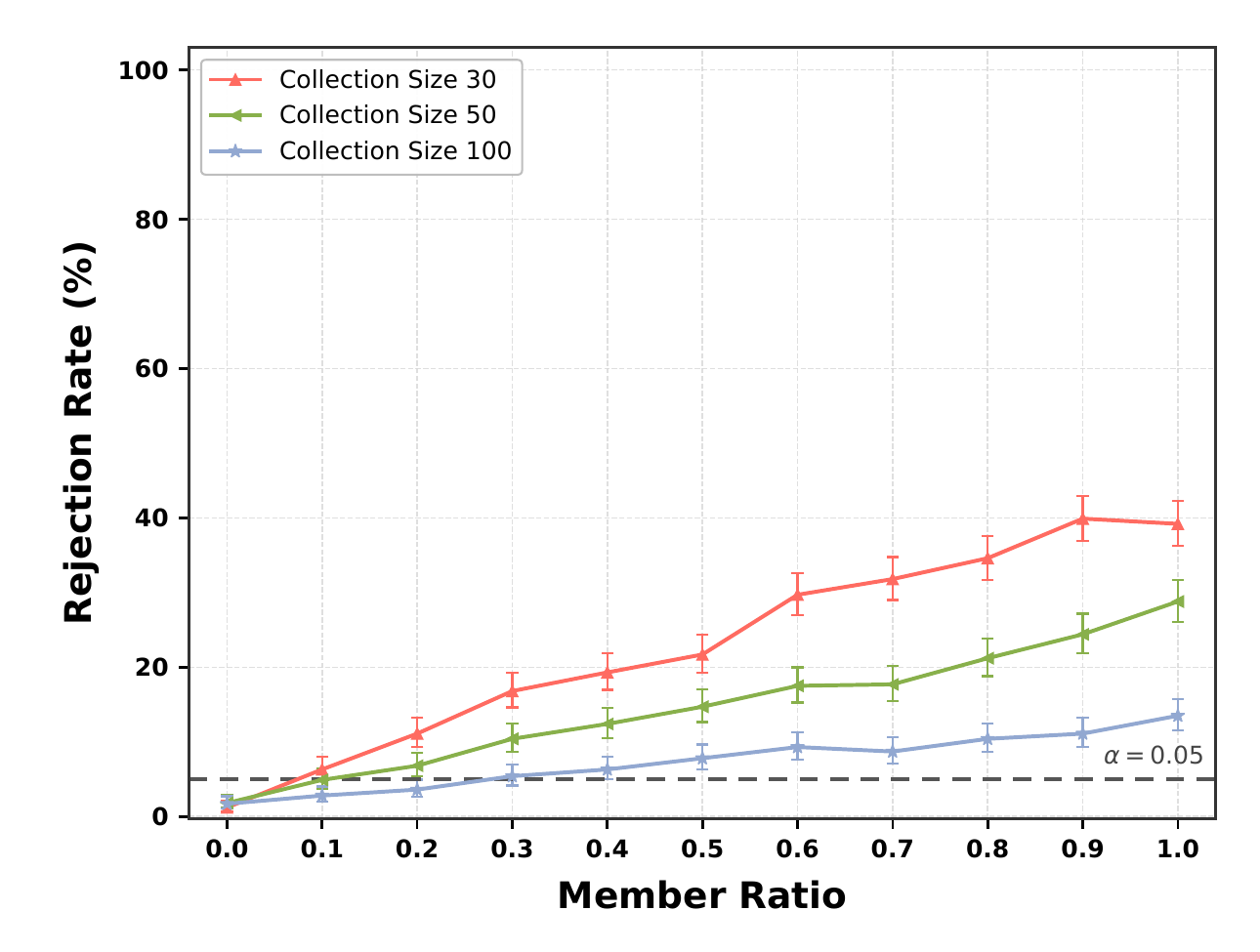}
        \caption{HCM}
        \label{fig:hcm_bonferroni}
    \end{subfigure}
    \caption{Bonferroni collection-level rejection rates on Pile--arXiv, Cosmopedia, and HCM. A member ratio of zero denotes entirely non-member collections. Error bars show pointwise $95\%$ Wilson confidence intervals. The dashed line marks the nominal significance level of $5\%$.}
    \label{fig:bonferroni_comparison}
\end{figure}

Bonferroni testing uses the data splits, target-collection sampling, and significance level specified in Section~\ref{app:evaluation_protocols}.
As shown in Figure~\ref{fig:bonferroni_comparison}, Bonferroni rejection rates generally increase with the member ratio, but the effect of collection size differs.
On Pile--arXiv, rejection rates increase with the member ratio, reaching $82.4\%$ and $94.0\%$ for entirely member collections of sizes $30$ and $50$, respectively. The size-$50$ collections have higher rejection rates throughout the positive member-ratio range, whereas no rejections occur for size-$100$ collections.
On Cosmopedia, the curves are close, with $m=100$ generally giving the highest rejection rate; entirely member collections yield rates of about $70$--$76\%$.
On HCM, larger collections yield lower rejection rates at positive member ratios, with the $m=100$ curve remaining below $15\%$.
For entirely non-member collections on Pile--arXiv, the rejection rate is $6.8\%$ at $m=30$, while no rejections occur at $m=50$ or $m=100$. Rejection rates are close to $5\%$ on Cosmopedia and lower on HCM.

At a fixed positive member ratio, increasing $m$ adds members but also reduces the threshold $\alpha/m$.
The balance between these effects depends on how often document $p$-values reach the adjusted threshold.
Empirical calibration further imposes $p_i\geq 1/(n_{\mathrm{cal}}+1)$, so rejection is possible on the rank grid only when
\begin{equation}
\label{eq:app_bonferroni_resolution}
n_{\mathrm{cal}}\geq\left\lceil\frac{m}{\alpha}\right\rceil-1.
\end{equation}
For Pile--arXiv, the $1{,}850$ calibration documents (Table~\ref{tab:audit_splits}) imply $p_i\geq 1/1{,}851>0.05/100$, making rejection impossible at $m=100$ and explaining the zero curve.

\section{Size and Distribution of the Calibration Collection}
\label{app:hcm_calibration_sensitivity}

We study two aspects of calibration for BH subset identification on HCM: the size of the calibration collection and its distribution relative to target non-members.
The size comparison examines how the number of calibration documents affects empirical $p$-value resolution and member identification.
The distribution comparison perturbs the calibration population through stratified sampling based on answer length, keeping calibration size fixed.
This comparison is motivated by Theorem~4.6, which bounds the additional FDR cost of departure from the conditional exchangeability in Assumption~4.2(i) by the total variation distance between the actual and ideal joint score distributions.

We use the $m=30$ setting and evaluation protocol of Section~\ref{app:evaluation_protocols}, keeping the fitted evidence map, document scores, evaluation pools, and target collections fixed across calibration conditions.
Each condition uses one calibration collection throughout; only document $p$-values and BH decisions are recomputed.

\subsection{Distribution of the Calibration Collection}
\label{app:hcm_calibration_composition}

We perturb the calibration distribution at fixed size $n_{\mathrm{cal}}=1{,}000$ by stratifying non-members according to whether their scored answers contain at most $120$ tokens (short) or more than $120$ tokens (long).
The three calibration collections contain $800/200$, $500/500$, and $200/800$ short/long answers.
The $500/500$ condition closely matches the answer-length distribution of the evaluation non-member pool, in which $51.58\%$ of answers are short.
The $800/200$ and $200/800$ conditions induce mismatches toward shorter and longer answers, respectively.

\begin{figure}[!htbp]
    \centering
    \captionsetup{skip=4pt}
    \captionsetup[subfigure]{skip=2pt}
    \begin{subfigure}[t]{0.49\linewidth}
        \centering
        \includegraphics[width=\linewidth]{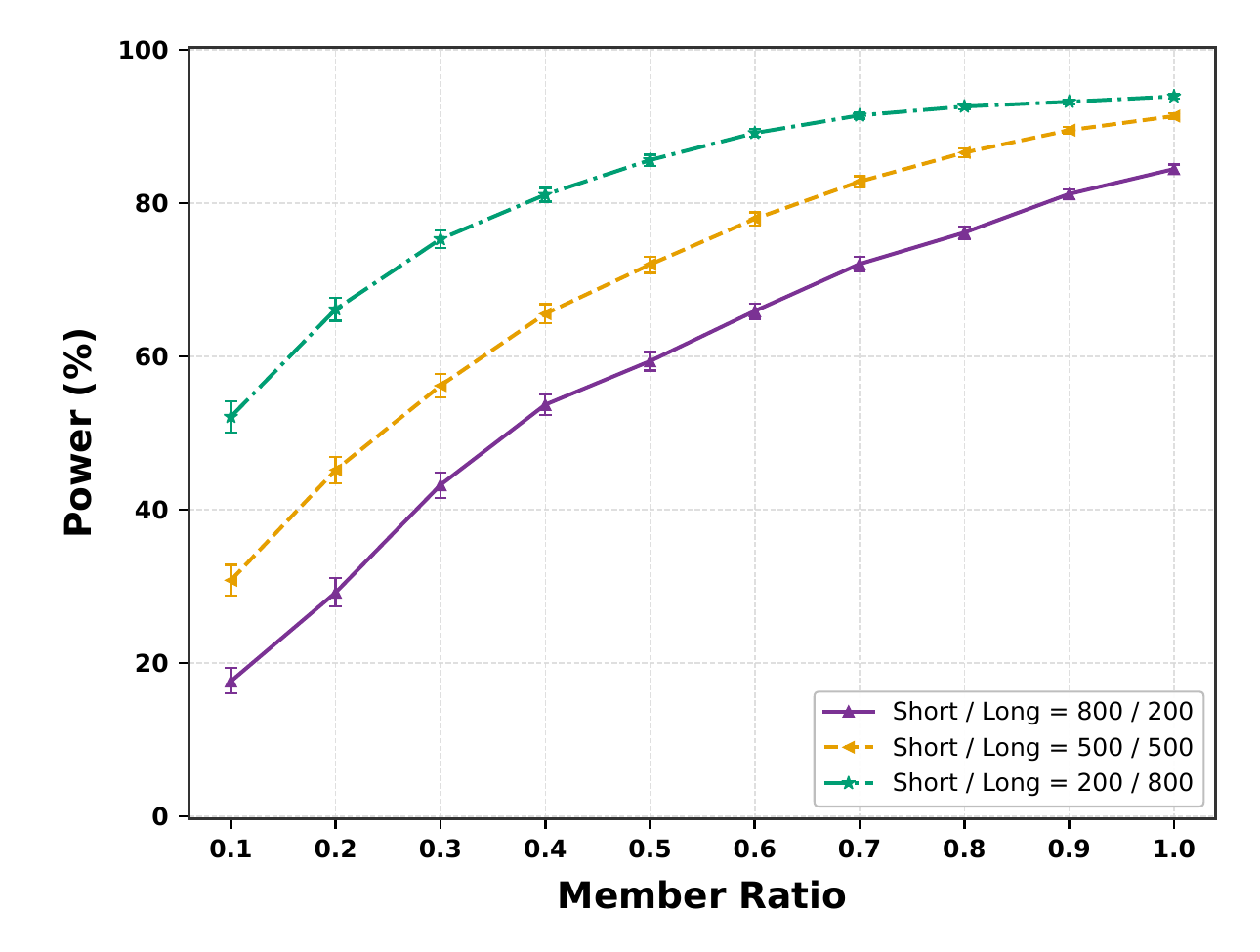}
        \caption{Power}
        \label{fig:hcm_calibration_composition_power}
    \end{subfigure}
    \hfill
    \begin{subfigure}[t]{0.49\linewidth}
        \centering
        \includegraphics[width=\linewidth]{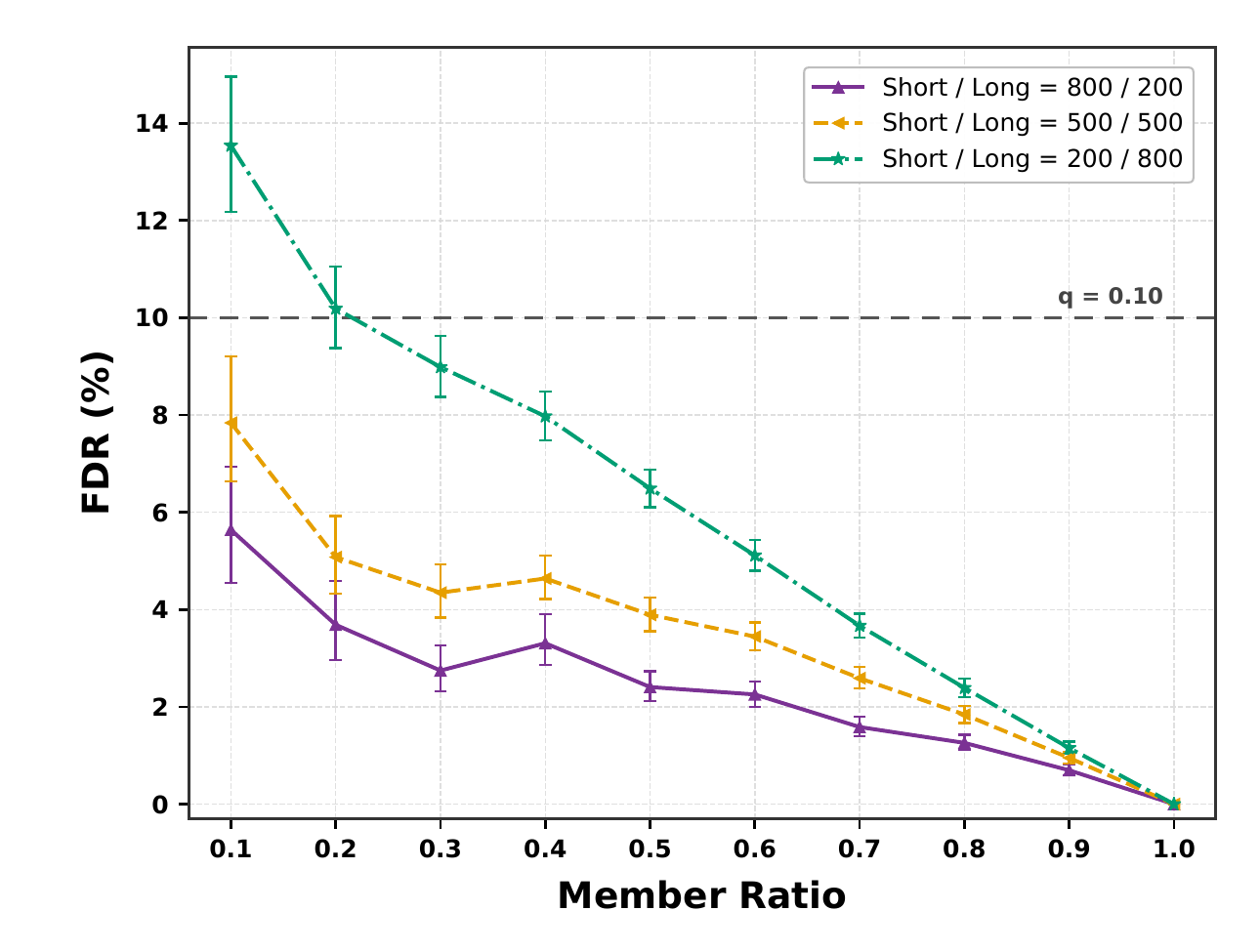}
        \caption{FDR}
        \label{fig:hcm_calibration_composition_fdr}
    \end{subfigure}
    \caption{BH subset identification on HCM with different calibration answer-length compositions ($n_{\mathrm{cal}}=1{,}000$). Short answers contain at most $120$ scored tokens. Error bars show BCa bootstrap confidence intervals; the dashed line marks the target FDR level.}
    \label{fig:hcm_calibration_composition}
\end{figure}

Figure~\ref{fig:hcm_calibration_composition} shows that increasing the proportion of long calibration answers raises both power and empirical FDR.
At a $10\%$ member ratio, power is approximately $18\%$, $31\%$, and $52\%$ for the three compositions, while empirical FDR is approximately $5.6\%$, $7.8\%$, and $13.5\%$, respectively.
The long-answer-heavy condition exceeds the nominal $10\%$ FDR level at this member ratio and lies close to that level at a $20\%$ member ratio.
The short-answer-heavy and balanced conditions remain below the nominal level throughout the displayed range.

The joint increase in power and FDR reflects a change in calibration of the same document evidence.
In Equation~(9), fewer calibration scores at least as large as a target score produce a smaller $p$-value, making BH selection more likely.
The differences are most pronounced when members are sparse, making calibration composition particularly relevant to audits seeking a small number of member documents.

\subsection{Size of the Calibration Collection}
\label{app:hcm_calibration_size}

\begin{figure}[!htbp]
    \centering
    \captionsetup{skip=4pt}
    \captionsetup[subfigure]{skip=2pt}
    \begin{subfigure}[t]{0.49\linewidth}
        \centering
        \includegraphics[width=\linewidth]{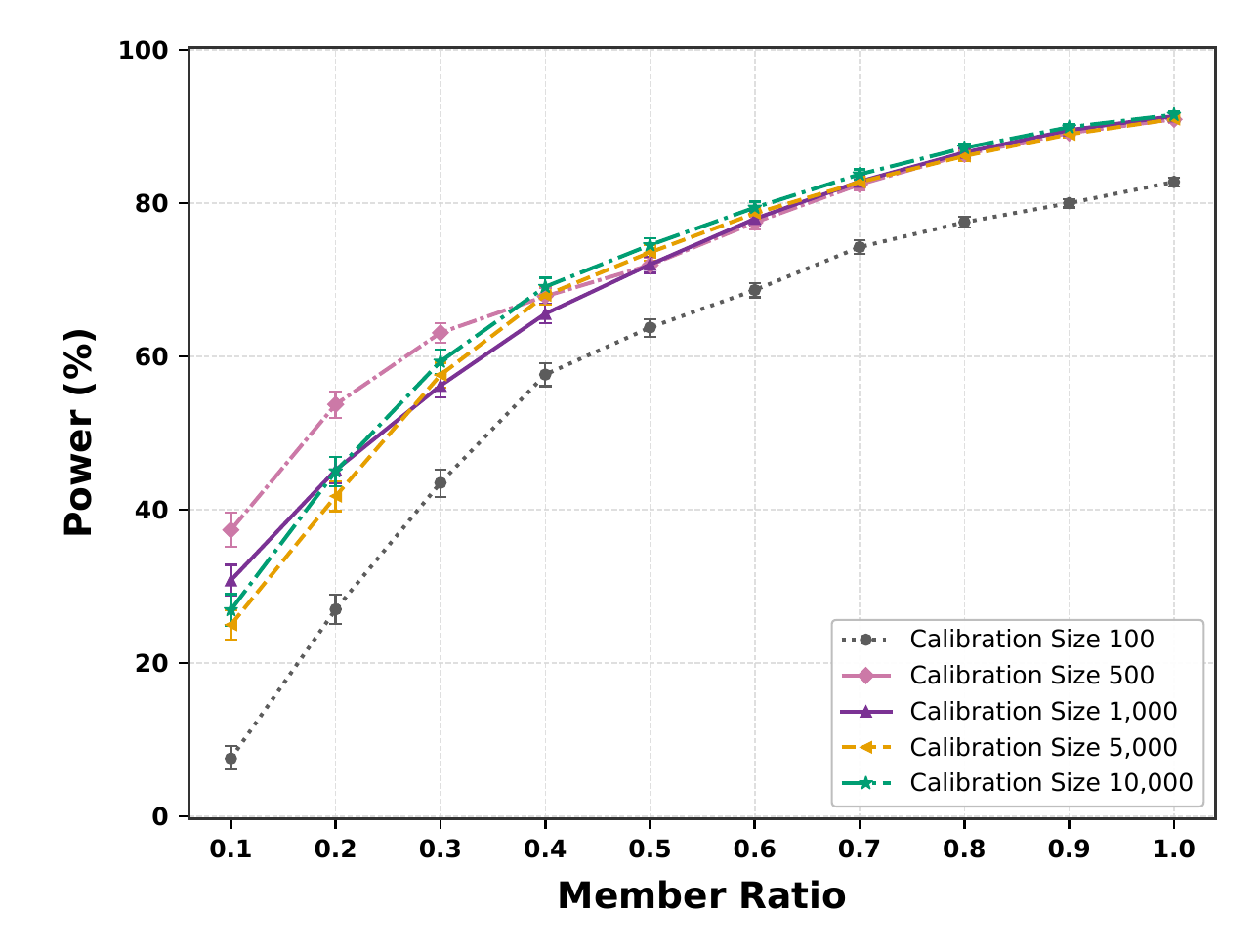}
        \caption{Power}
        \label{fig:hcm_calibration_size_power}
    \end{subfigure}
    \hfill
    \begin{subfigure}[t]{0.49\linewidth}
        \centering
        \includegraphics[width=\linewidth]{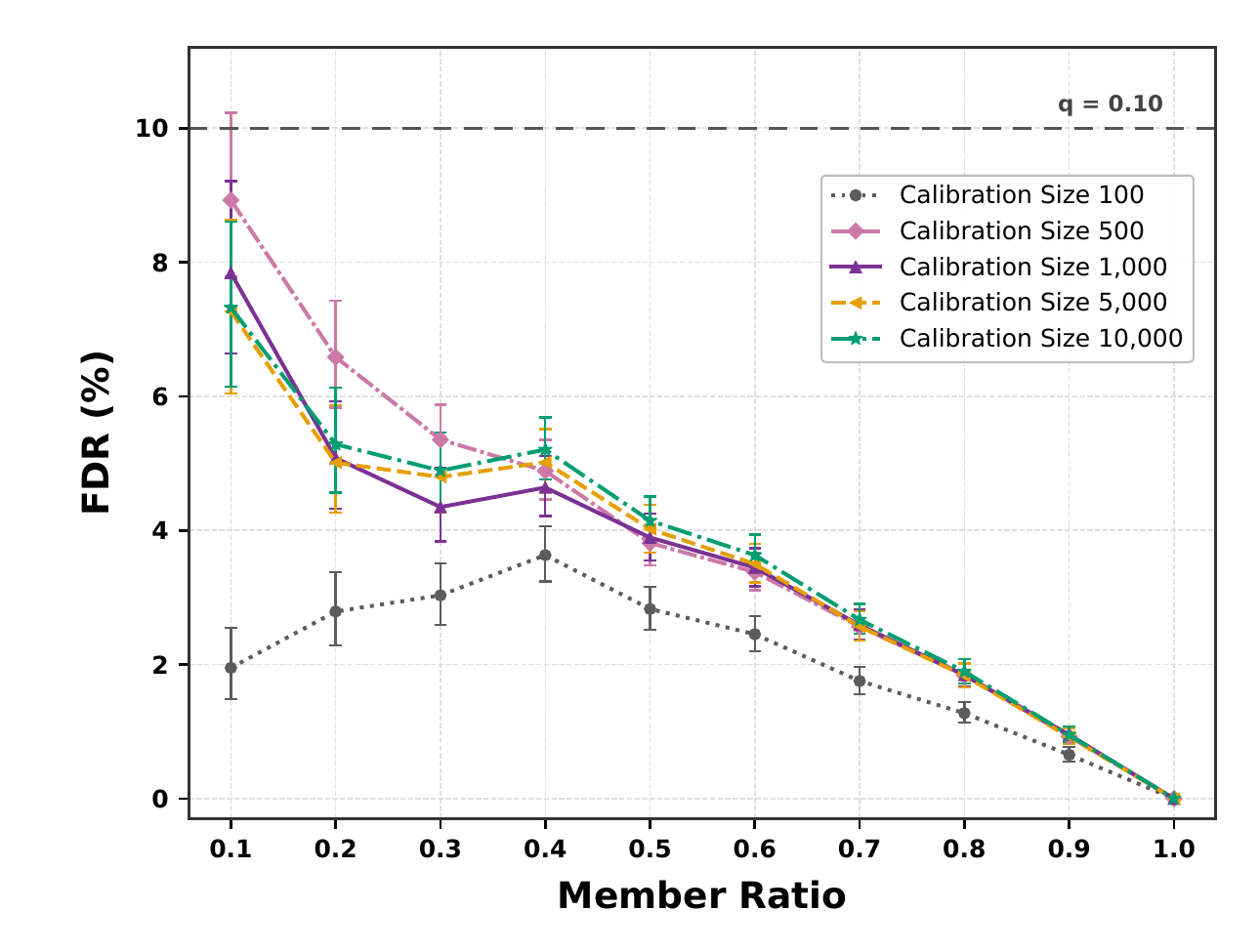}
        \caption{FDR}
        \label{fig:hcm_calibration_size_fdr}
    \end{subfigure}
    \caption{BH subset identification on HCM with different calibration collection sizes. Error bars show BCa bootstrap confidence intervals; the dashed line marks the target FDR level.}
    \label{fig:hcm_calibration_size}
\end{figure}

We next compare calibration sizes $n_{\mathrm{cal}}\in\{100,500,1{,}000,5{,}000,10{,}000\}$.
Calibration documents are drawn from the original non-member pool without deliberately shifting its answer-length composition; the largest condition uses the complete pool.
Figure~\ref{fig:hcm_calibration_size} shows similar power and empirical FDR for the $1{,}000$, $5{,}000$, and $10{,}000$ conditions over much of the member-ratio range.
For example, at a $50\%$ member ratio, power is approximately $72$--$75\%$ across these three sizes.
The $100$-document condition has lower power throughout the displayed range, accompanied by lower empirical FDR.
All five conditions have empirical FDR point estimates below $10\%$.

The lower power at $n_{\mathrm{cal}}=100$ is consistent with limited empirical rank resolution.
The minimum $p$-value, $1/101\approx0.0099$, exceeds the first two BH thresholds, $0.10/30\approx0.0033$ and $2(0.10)/30\approx0.0067$.
BH therefore cannot return a discovery set of size one or two, restricting identification when members are sparse.
At $n_{\mathrm{cal}}=1{,}000$, the minimum $p$-value is approximately $0.0010$, below the first BH threshold.

The $500$-document collection gives higher power and FDR than larger collections at the lowest member ratios, showing that realized calibration scores also affect performance.
For these fixed HCM calibration collections, increasing size from $1{,}000$ to $10{,}000$ has a smaller effect than changing length composition at size $1{,}000$.

\FloatBarrier

%% file: booktection_results.tex
\begin{figure}[!htbp]
    \centering
    \renewcommand{\baselinestretch}{1.1}\selectfont
    \captionsetup{font=small,skip=7pt,justification=raggedright,singlelinecheck=false}
    \newsavebox{\booktectionresultstablebox}
    \sbox{\booktectionresultstablebox}{%
    \footnotesize
    \renewcommand{\arraystretch}{1.15}
    \setlength{\tabcolsep}{4pt}
    \begin{tabular}{@{}lccc@{}}
        \toprule
        \multirow{2}{*}{\textbf{Method}} & \multicolumn{2}{c}{\textbf{TPR@FPR}}
        & \multirow{2}{*}{\textbf{AUC}} \\
        \cmidrule(lr){2-3}
        & $1\%^{\dagger}$ & $5\%$ & \\
        \midrule
        PPL & 7.06 & 41.18 & 81.66 \\
        Min-$K$ $20\%$ & 7.06 & 43.53 & 84.03 \\
        Mean-evidence & 57.65 & 84.71 & 95.56 \\
        Mann--Whitney U & 48.24 & 85.88 & 94.60 \\
        Fisher combination & \underline{61.18} & \textbf{90.59} & \underline{96.15} \\
        Cauchy (OURS) & \textbf{80.00} & \underline{89.41} & \textbf{96.41} \\
        \bottomrule
    \end{tabular}%
    }
    \newsavebox{\booktectionresultsplotbox}
    \sbox{\booktectionresultsplotbox}{\includegraphics[width=1.1\wd\booktectionresultstablebox]{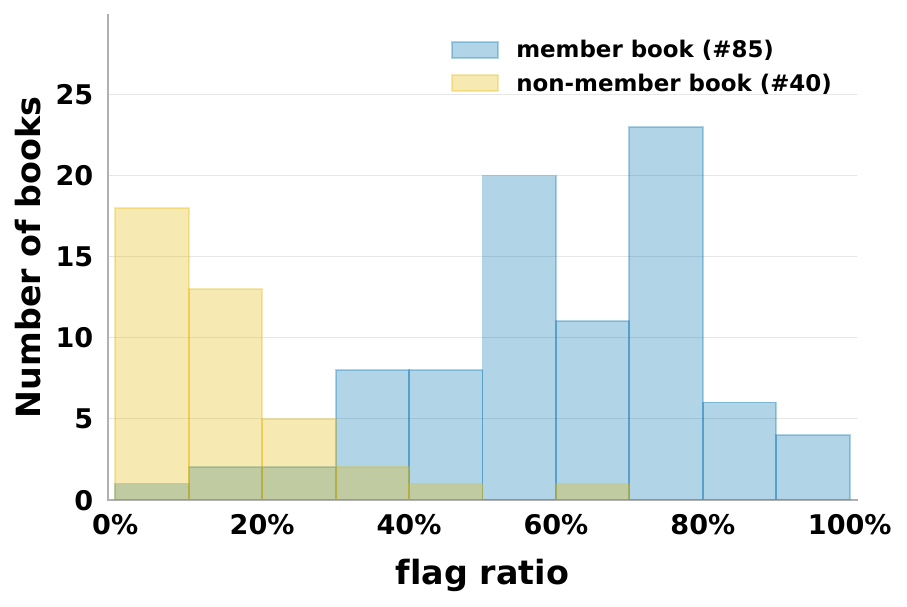}}
    \newlength{\booktectionresultsheight}
    \setlength{\booktectionresultsheight}{\dimexpr\ht\booktectionresultsplotbox+\dp\booktectionresultsplotbox\relax}
    \begin{minipage}[t]{0.49\linewidth}
        \vspace{0pt}
        \begin{minipage}[c][\booktectionresultsheight][c]{\linewidth}
            \centering
            \resizebox*{!}{0.96\booktectionresultsheight}{\usebox{\booktectionresultstablebox}}
        \end{minipage}
        \captionof{table}{Book-level discrimination (\%) on BookTection with Pythia-6.9B. Best in bold; second best underlined.}
        \label{tab:booktection_auc}
    \end{minipage}
    \hfill
    \begin{minipage}[t]{0.49\linewidth}
        \vspace{0pt}
        \begin{minipage}[c][\booktectionresultsheight][c]{\linewidth}
            \centering
            \usebox{\booktectionresultsplotbox}
        \end{minipage}
        \captionsetup{justification=justified,singlelinecheck=true}
        \captionof{figure}{Flagged-passage proportions for candidate-member and non-member books in the BookTection Medium setting.}
        \label{fig:booktection_flagged_chunks}
    \end{minipage}
    \par\vspace{5pt}
    \begin{minipage}{\linewidth}
        \footnotesize\raggedright
        \textsuperscript{\textdagger} With $40$ non-member books, empirical FPR increases in steps of $2.5\%$; TPR@$1\%$ FPR therefore reports the maximum TPR at zero empirical FPR.
    \end{minipage}
\end{figure}

%% file: booktection_summary.tex
\begin{table}[!ht]
\centering
\renewcommand{\baselinestretch}{1.1}\small
\caption{Ten BookTection books with the highest proportions of passages satisfying $p_{ij}^{\mathrm{ck}}<0.1$ in the Medium setting. The bounded Cauchy-transformed score $u(d)$ is computed using Equation~(10); smaller values indicate stronger aggregated membership evidence.}
\label{tab:book-examples-large}
\setlength{\tabcolsep}{4pt}
\renewcommand{\arraystretch}{0.6}
\begin{tabular*}{\textwidth}{@{\extracolsep{\fill}}lrr@{}}
\toprule
\textbf{Book title} & \textbf{Flag ratio (\%)} & \textbf{\boldmath$u(d)$ ($\times 10^{-2}$)} \\
\midrule
Alice's Adventures in Wonderland & 100.0 & $0.149$ \\
The Great Gatsby & 94.1 & $0.188$ \\
The Adventures of Sherlock Holmes & 94.1 & $0.258$ \\
A Game of Thrones & 91.2 & $0.314$ \\
The Picture of Dorian Gray & 88.2 & $0.285$ \\
Emma & 87.9 & $0.341$ \\
Lord of the Rings: The Fellowship of the Ring & 85.3 & $0.339$ \\
Oliver Twist & 82.4 & $0.432$ \\
The Boy in the Striped Pajamas & 82.4 & $0.602$ \\
The Silmarillion & 81.8 & $0.361$ \\
\bottomrule
\end{tabular*}
\end{table}